\documentclass[a4paper,11pt]{amsart}
\usepackage{amsfonts}
\usepackage{fontenc}
\usepackage [utf8]{inputenc}
\usepackage{ xcolor}
\usepackage[left=3cm,right=3cm,top=2.8cm,bottom=3.8cm]{geometry}
\usepackage{xcolor,colortbl}
\definecolor{green}{rgb}{0.1,0.1,0.1}

\usepackage{graphicx}
\usepackage{graphics}
\usepackage{graphicx}
\usepackage{array,multirow,makecell}
\usepackage{algorithm}
\usepackage[noend]{algpseudocode}
\renewcommand{\algorithmicrequire}{\textbf{Input: }}
\renewcommand{\algorithmicensure}{\textbf{Output: }}
\usepackage[colorlinks=true,citecolor=black,linkcolor=black,urlcolor=blue]{hyperref}
\usepackage{color}
\usepackage{graphicx} %%For loading graphic files
\usepackage{amssymb, amsmath}
\usepackage{setspace}
\usepackage{hyperref}
\usepackage{amsfonts}

\usepackage{hyperref}
\usepackage{tikz}
\usepackage{pgfplots}
\usepackage{pstricks-add}
\usepackage{algorithm,algpseudocode}

\def\dsum{\displaystyle\sum}

\begin{document}
\title[A new hybrid GA for PSP on the 2D triangular lattice]{A new hybrid genetic algorithm for protein structure prediction on the 2D triangular lattice}

\author[Nabil Boumedine and Sadek Bouroubi]{Nabil Boumedine$^1$,  Sadek BOUROUBI$^2$\vspace{2mm} \\ \MakeLowercase{\texttt{nabil.doukou@gmail.com}$^1$, \texttt{sbouroubi@usthb.dz}$^2$}  \vspace{3mm}\\ $^{1,2}$ LIFORCE L\MakeLowercase{aboratory}, \vspace{1mm}\\
USTHB, F\MakeLowercase{aculty of Mathematics}, D\MakeLowercase{epartment of} O\MakeLowercase{perations} R\MakeLowercase{esearch},\vspace{1mm} \\ P.B. 32 E\MakeLowercase{l-Alia}, 16111, B\MakeLowercase{ab Ezzouar}, A\MakeLowercase{lgiers}, A\MakeLowercase{lgeria}.
}

%%%%%%%%%%%%%%%%%%%%%%%%%%%%%%%%%%%%%%%%%%%%%%%%%%%%%%%%%%%%%%%%%%%

\begin{abstract}
The flawless functioning of a protein is essentially linked to its own three-dimensional structure. Therefore, the prediction of a protein structure from its amino acid sequence is a fundamental problem in many fields  that draws researchers attention. This problem can be \mbox{formulated} as a combinatorial optimization problem based on simplified lattice \mbox{models} such as the hydrophobic-polar model. In this  paper, we propose a  new hybrid algorithm combining three different well-known heuristic algorithms: genetic algorithm, tabu search strategy and  local search algorithm in order to solve the PSP problem. Regarding the assessment of suggested algorithm, an experimental study is included, where we considered the quality of the produced solution as the main quality criterion. Furthermore, we compared the suggested algorithm with state-of-the-art algorithms using a selection of well-studied benchmark instances.
\vspace{3mm}

\noindent\textsc{Keywords and phrases.} Protein Structure Prediction, 2D triangular
Lattice, HP Model, Genetic Algorithm, Local Search, Tabu Search Strategy,  Minimal Energy Conformation.\vspace{-2.5em}

\end{abstract}

%%%%%%%%%%%%%%%%%%%%%%%%%%%%%%%%%%%%%%%%%%%%%%%%%%%%%%%%%%%%%%%%%%%

\maketitle

\section{Introduction}
In Molecular biology, the  three dimensional structure of proteins is the most crucial indicator that determines their biological activity. The prediction of the tertiary structure of proteins, also called a conformation, is a fundamental problem in the fields of biology, physics, etc. This problem is famous for protein structure prediction and denoted by PSP. A small modification in  the original conformation of any given protein or an error during its folding causes many serious diseases, such as Alzheimer and cow mad. The ideal solution to treat these diseases is to predict the tertiaries structures of these proteins from their primary structure \cite{dill_theory_1985}.
The PSP is one of the hardest problems in computational biology, molecular biology, biochemistry and physics. Furthermore, The most efficient algorithm  for  feasible solution determination runs in an  an considerable time required; whereas the correct functioning of a protein  depends essentially on its minimal energy conformation.  Among the existing models in literature, the most studied in PSP problem is the H-P model (Hydrophobic Polar, H-P) \cite{lau_lattice_1989}.In this model the free energy of a conformation is inversely  proportional to the number of  hydrophobic  non-local bonds of H-H type (topological contacts H-H). This type  of bonds occurs if two non-consecutive hydrophobic monomers occupy adjacent grid points in the lattice.  Besides, each occurrence of this bond type reduces the value of global energy with one unit \cite{lin_efficient_2009}. The PSP is an optimization problem where  the aim is to find a confirmation $c^{*}$ of a given protein sequence that minimizes the overall induced energy in all possible set of  conformations $C$, i.e., a conformation $c^{*}$ such that $E(c^{*})=min\{E(c)/c\in C\}$\cite{shmygelska_ant_2005},   where $E(c)$ represent the energy function explained later in Section 2.2.
 As we mentioned above, the H-H bonds  reduces the induced energy. Hence, finding a minimal energy conformation (i.e., optimal conformation) amounts to find a conformation that maximizes the number of H-H contacts \cite{lin_efficient_2009}. As one may expect,  the resolution of this problem is quite difficult due to the exponential exploration of the NP-hard problem solving when the chain size is large enough. The problem  is proven to be NP-hard even when restricted  to its two dimensional representation \cite{berger_protein_1998}.Hence, it is clearly impossible to enumerate all the possible conformations when the chain of amino acids is  considerably large. Several metaheuristics  were proposed to solve the PSP problem. In the two-dimensional square lattice, the first  genetic algorithm has been introduced by Unger and Moult \cite{unger_genetic_1993}, and than followed by other versions, see \cite{dandekar_folding_1994, konig_improving_1999, krasnogor_protein_1999}. Similarly, an ant colony optimization (ACO) algorithm has been used by  Shmygelska et al. \cite{shmygelska_ant_2002, shmygelska_improved_2003, shmygelska_ant_2005}. Moreover, the use of memetic algorithms was proposed in \cite{krasnogor_multimeme_2002, pelta_multimeme_2005}. Particle Swarm Optimization algorithms has been applied in \cite{bautu_protein_2010}. Jiang, T. et al. proposed a hybrid approach combining tabu search and genetic algorithm \cite{jiang_protein_2003}. The Immune algorithms are introduced by Cutello et al. in \cite{cutello_immune_2005,cutello_immune_2007}. A Clustered memetic algorithm with local heuristics have been introduced by Islam et al. in \cite{islam_clustered_2013}.

This paper  is organized as follows: in the next section, we present the two dimensional triangular lattice using the H-P model, that we are  interested in, where we present a corresponding 0-1 mathematical program while focusing on: the objective function and the encoding solutions. In Section 3 we present a selection of pertinent algorithms designed to solve the PSP problem in 2D triangular lattice model. In Section 3, we present a detailed description of the suggested  hybrid algorithm. In Section 4, we present the experimental study and the obtained results, where we compared our approach some existing approaches. Finally, we give, in Section 6, our main conclusions of this study.
\section{Hydrophobic-polar model in a 2D triangular  lattice Simplified}
 In H-P model, the twenty  amino acids are represented by a mean of two  letters H and P,  in reference to their hydrophobicity chosen among the tow following options : hydrophobic or hydrophilic  \cite{lau_lattice_1989}. For  any given sequence of $n$ amino acids, the H-P model consists to transform this sequence into a chain  $s = (s_1, s_2, \dots, s_n)$ such that each element of the sequence $s_{i}\in\{H,P\},\ i=\overline{1,n}$ represents the hydrophobicity of the corresponding amino acids in protein sequence:
 $$s_{i}=\left\{
  \begin{array}{ll}
    H & \hbox{if the amino acid $i$ is of hydrophobic type,} \vspace{0.1 cm}\\
    P & \hbox{if the amino acid $i$ is of polar type.}
  \end{array}
\right.$$
As it is shown in Figure 1, each vertex of the two dimensional triangular lattice has six neighbors. Hence, each monomer different from the first and the last element of the chain, i.e., monomers of rank \mbox{$i=\overline{2,n-1}$}, occupying any given position in the lattice can be at most in four topological contacts, in other words, it can form at most four bonds with its neighbors. Whereas maximal number of possible contacts that can occur in a monomer located either in the first or the last position is five  \cite{agarwala_local_1997}. The symbols $1, 2, 3, 4, 5, 6$ are used to encode the following movement directions on the two dimensional triangular lattice:  Right-Up, Up, Left-Up, Left-Down, Down and  Right-Down, respectively.

\begin{figure}[H]
\begin{center}
\psset{xunit=1.0cm,yunit=1.0cm,algebraic=true,dimen=middle,dotstyle=o,dotsize=5pt 0,linewidth=0.8pt,arrowsize=3pt 2,arrowinset=0.25}
\begin{pspicture*}(7.661386664822013,6.391342379551787)(11.371825732930434,10.614927276228414)
\psaxes[labelFontSize=\scriptstyle,xAxis=true,yAxis=true,Dx=1.,Dy=1.,ticksize=-2pt 0,subticks=2]{->}(0,0)(7.661386664822013,6.391342379551787)(11.371825732930434,10.614927276228414)
\psline[linestyle=dashed,dash=1pt 1pt]{->}(9.526279441628825,8.5)(10.392304845413264,9.)
\psline[linestyle=dashed,dash=1pt 1pt]{->}(9.526279441628825,8.5)(9.526279441628825,9.5)
\psline[linestyle=dashed,dash=1pt 1pt]{->}(9.526279441628825,8.5)(8.660254037844386,9.)
\psline[linestyle=dashed,dash=1pt 1pt]{->}(9.526279441628825,8.5)(8.660254037844386,8.)
\psline[linestyle=dashed,dash=1pt 1pt]{->}(9.526279441628825,8.5)(9.526279441628825,7.5)
\psline[linestyle=dashed,dash=1pt 1pt]{->}(9.526279441628825,8.5)(10.392304845413264,8.)
\rput[tl](10.482816988164814,9.259169301139922){\textcolor[rgb]{1.00,0.00,0.00}{1}}
\rput[tl](9.429327911535822,9.888693261564564){\textcolor[rgb]{1.00,0.00,0.00}{2}}
\rput[tl](8.350143979379295,9.259169301139922){\textcolor[rgb]{1.00,0.00,0.00}{3}}
\rput[tl](8.350143979379295,8.057934805227596){\textcolor[rgb]{1.00,0.00,0.00}{4}}
\rput[tl](9.429327911535822,7.293512853283389){\textcolor[rgb]{1.00,0.00,0.00}{5}}
\rput[tl](10.482816988164814,8.057934805227596){\textcolor[rgb]{0.98,0.00,0.00}{6}}
\psline[linewidth=0.4pt,linestyle=dashed,dash=1pt 1pt,linecolor=lightgray](7.794228634059948,9.5)(9.52627944162883,10.5)
\psline[linewidth=0.4pt,linestyle=dashed,dash=1pt 1pt,linecolor=lightgray](9.52627944162883,10.5)(11.258330249197705,9.5)
\psline[linewidth=0.4pt,linestyle=dashed,dash=1pt 1pt,linecolor=lightgray](11.258330249197705,9.5)(11.258330249197705,7.5)
\psline[linewidth=0.4pt,linestyle=dashed,dash=1pt 1pt,linecolor=lightgray](9.52627944162883,6.5)(7.794228634059941,7.5)
\psline[linewidth=0.4pt,linestyle=dashed,dash=1pt 1pt,linecolor=lightgray](11.258330249197705,7.5)(9.52627944162883,6.5)
\psline[linewidth=0.4pt,linestyle=dashed,dash=1pt 1pt,linecolor=lightgray](7.794228634059941,7.5)(7.794228634059948,9.5)
\psline[linewidth=0.4pt,linestyle=dashed,dash=1pt 1pt,linecolor=lightgray](8.660254037844386,10.)(8.660254037844396,7.)
\psline[linewidth=0.4pt,linestyle=dashed,dash=1pt 1pt,linecolor=lightgray](9.52627944162883,10.5)(9.52627944162883,9.5)
\psline[linewidth=0.4pt,linestyle=dashed,dash=1pt 1pt,linecolor=lightgray](9.526279441628816,7.5)(9.52627944162883,6.5)
\psline[linewidth=0.4pt,linestyle=dashed,dash=1pt 1pt,linecolor=lightgray](10.392304845413259,10.)(10.392304845413282,7.)
\psline[linewidth=0.4pt,linestyle=dashed,dash=1pt 1pt,linecolor=lightgray](7.794228634059948,8.5)(10.392304845413259,10.)
\psline[linewidth=0.4pt,linestyle=dashed,dash=1pt 1pt,linecolor=lightgray](7.794228634059941,7.5)(8.660254037844393,8.)
\psline[linewidth=0.4pt,linestyle=dashed,dash=1pt 1pt,linecolor=lightgray](10.392304845413266,9.)(11.258330249197705,9.5)
\psline[linewidth=0.4pt,linestyle=dashed,dash=1pt 1pt,linecolor=lightgray](8.660254037844396,7.)(11.258330249197705,8.5)
\psline[linewidth=0.4pt,linestyle=dashed,dash=1pt 1pt,linecolor=lightgray](7.794228634059948,9.5)(8.660254037844389,9.)
\psline[linewidth=0.4pt,linestyle=dashed,dash=1pt 1pt,linecolor=lightgray](10.392304845413273,8.)(11.258330249197705,7.5)
\psline[linewidth=0.4pt,linestyle=dashed,dash=1pt 1pt,linecolor=lightgray](7.794228634059948,8.5)(10.392304845413282,7.)
\psline[linewidth=0.4pt,linestyle=dashed,dash=1pt 1pt,linecolor=lightgray](8.660254037844386,10.)(11.258330249197705,8.5)
\begin{scriptsize}
\psdots[dotstyle=*,linecolor=red](9.526279441628825,8.5)
%\rput[bl](2.3009863940866544,11.293858680180172){$Curve$}
\end{scriptsize}
\end{pspicture*}
\caption{Illustration of the six neighbors of a node in the 2D triangular lattice model.}
\label{figure1}
\end{center}
\end{figure}
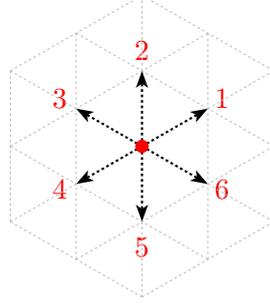
\begin{figure}[H]
\begin{center}
\newrgbcolor{eqeqeq}{0.8784313725490196 0.8784313725490196 0.8784313725490196}
\psset{xunit=1.0cm,yunit=1.0cm,algebraic=true,dimen=middle,dotstyle=o,dotsize=5pt 0,linewidth=0.8pt,arrowsize=3pt 2,arrowinset=0.25}
\begin{pspicture*}(2.3917106533551533,2.2861037342378885)(8.839568481735444,9.229950626339718)
\psaxes[labelFontSize=\scriptstyle,xAxis=true,yAxis=true,Dx=1.,Dy=1.,ticksize=-2pt 0,subticks=2]{->}(0,0)(2.3917106533551533,2.2861037342378885)(8.839568481735444,9.229950626339718)
\rput[tl](-0.002719309438589181,0.19952905237477697){1}
\psline[linewidth=1.2pt,linecolor=blue](3.4641016151377544,6.)(3.4641016151377544,7.)
\psline[linewidth=1.2pt,linecolor=blue](3.4641016151377544,7.)(4.330127018922193,6.5)
\psline[linewidth=1.2pt,linecolor=blue](4.330127018922193,6.5)(4.330127018922193,7.5)
\psline[linewidth=1.2pt,linecolor=blue](4.330127018922193,7.5)(5.196152422706632,7.)
\psline[linewidth=1.2pt,linecolor=blue](5.196152422706632,7.)(5.196152422706632,6.)
\psline[linewidth=1.2pt,linecolor=blue](5.196152422706632,6.)(4.330127018922193,5.5)
\psline[linewidth=1.2pt,linecolor=blue](4.330127018922193,5.5)(4.330127018922193,4.5)
\psline[linewidth=1.2pt,linecolor=blue](4.330127018922193,4.5)(5.196152422706632,5.)
\psline[linewidth=1.2pt,linecolor=blue](5.196152422706632,5.)(5.196152422706632,4.)
\psline[linewidth=1.2pt,linecolor=blue](5.196152422706632,4.)(6.06217782649107,3.5)
\psline[linewidth=1.2pt,linecolor=blue](6.06217782649107,3.5)(6.06217782649107,4.5)
\psline[linewidth=1.2pt,linecolor=blue](6.06217782649107,4.5)(6.928203230275509,4.)
\psline[linewidth=1.2pt,linecolor=blue](6.928203230275509,4.)(6.928203230275509,5.)
\psline[linewidth=1.2pt,linecolor=blue](6.928203230275509,5.)(6.06217782649107,5.5)
\psline[linewidth=1.2pt,linecolor=blue](6.06217782649107,5.5)(6.06217782649107,6.5)
\psline[linewidth=1.2pt,linecolor=blue](6.06217782649107,6.5)(6.928203230275509,7.)
\psline[linewidth=1.2pt,linecolor=blue](6.928203230275509,7.)(6.928203230275509,6.)
\psline[linewidth=1.2pt,linecolor=blue](6.928203230275509,6.)(7.794228634059947,6.5)
\psline[linewidth=1.2pt,linecolor=blue](7.794228634059947,6.5)(7.794228634059947,5.5)
\psline[linewidth=0.4pt,linestyle=dashed,dash=2pt 2pt,linecolor=eqeqeq](2.598076211353316,7.5)(5.196152422706632,9.)
\psline[linewidth=0.4pt,linestyle=dashed,dash=2pt 2pt,linecolor=lightgray](2.598076211353316,7.5)(5.196152422706632,9.)
\psline[linewidth=0.4pt,linestyle=dashed,dash=2pt 2pt,linecolor=lightgray](5.196152422706632,9.)(8.660254037844386,7.)
\psline[linewidth=0.4pt,linestyle=dashed,dash=2pt 2pt,linecolor=lightgray](8.660254037844386,7.)(8.660254037844386,4.)
\psline[linewidth=0.4pt,linestyle=dashed,dash=2pt 2pt,linecolor=lightgray](8.660254037844386,4.)(6.06217782649107,2.5)
\psline[linewidth=0.4pt,linestyle=dashed,dash=2pt 2pt,linecolor=lightgray](6.06217782649107,2.5)(2.598076211353316,4.5)
\psline[linewidth=0.4pt,linestyle=dashed,dash=2pt 2pt,linecolor=lightgray](2.598076211353316,4.5)(2.598076211353316,7.5)
\psline[linewidth=0.4pt,linestyle=dashed,dash=2pt 2pt,linecolor=lightgray](2.598076211353316,7.5)(3.4641016151377544,7.)
\psline[linewidth=0.4pt,linestyle=dashed,dash=2pt 2pt,linecolor=lightgray](6.928203230275509,5.)(8.660254037844386,4.)
\psline[linewidth=0.4pt,linestyle=dashed,dash=2pt 2pt,linecolor=lightgray](3.4641016151377544,8.)(3.4641016151377544,7.)
\psline[linewidth=0.4pt,linestyle=dashed,dash=2pt 2pt,linecolor=lightgray](3.4641016151377544,6.)(3.464101615137755,4.)
\psline[linewidth=0.4pt,linestyle=dashed,dash=2pt 2pt,linecolor=lightgray](4.330127018922193,3.5)(4.330127018922193,4.5)
\psline[linewidth=0.4pt,linestyle=dashed,dash=2pt 2pt,linecolor=lightgray](4.330127018922193,5.5)(4.330127018922193,6.5)
\psline[linewidth=0.4pt,linestyle=dashed,dash=2pt 2pt,linecolor=lightgray](4.330127018922193,7.5)(4.330127018922193,8.5)
\psline[linewidth=0.4pt,linestyle=dashed,dash=2pt 2pt,linecolor=lightgray](5.196152422706632,9.)(5.196152422706632,7.)
\psline[linewidth=0.4pt,linestyle=dashed,dash=2pt 2pt,linecolor=lightgray](5.196152422706632,6.)(5.196152422706632,5.)
\psline[linewidth=0.4pt,linestyle=dashed,dash=2pt 2pt,linecolor=lightgray](5.196152422706632,4.)(5.196152422706632,3.)
\psline[linewidth=0.4pt,linestyle=dashed,dash=2pt 2pt,linecolor=lightgray](6.06217782649107,2.5)(6.06217782649107,3.5)
\psline[linewidth=0.4pt,linestyle=dashed,dash=2pt 2pt,linecolor=lightgray](6.06217782649107,4.5)(6.06217782649107,5.5)
\psline[linewidth=0.4pt,linestyle=dashed,dash=2pt 2pt,linecolor=lightgray](6.06217782649107,6.5)(6.06217782649107,8.5)
\psline[linewidth=0.4pt,linestyle=dashed,dash=2pt 2pt,linecolor=lightgray](6.928203230275508,8.)(6.928203230275509,7.)
\psline[linewidth=0.4pt,linestyle=dashed,dash=2pt 2pt,linecolor=lightgray](6.928203230275509,6.)(6.928203230275509,5.)
\psline[linewidth=0.4pt,linestyle=dashed,dash=2pt 2pt,linecolor=lightgray](6.928203230275509,4.)(6.928203230275509,3.)
\psline[linewidth=0.4pt,linestyle=dashed,dash=2pt 2pt,linecolor=lightgray](7.794228634059947,3.5)(7.794228634059947,5.5)
\psline[linewidth=0.4pt,linestyle=dashed,dash=2pt 2pt,linecolor=lightgray](7.794228634059947,6.5)(7.794228634059945,7.5)
\psline[linewidth=0.4pt,linestyle=dashed,dash=2pt 2pt,linecolor=lightgray](8.660254037844386,7.)(8.660254037844386,4.)
\psline[linewidth=0.4pt,linestyle=dashed,dash=2pt 2pt,linecolor=lightgray](4.330127018922193,8.5)(8.660254037844386,6.)
\psline[linewidth=0.4pt,linestyle=dashed,dash=2pt 2pt,linecolor=lightgray](3.4641016151377544,8.)(4.330127018922193,7.5)
\psline[linewidth=0.4pt,linestyle=dashed,dash=2pt 2pt,linecolor=lightgray](5.196152422706632,7.)(6.06217782649107,6.5)
\psline[linewidth=0.4pt,linestyle=dashed,dash=2pt 2pt,linecolor=lightgray](6.06217782649107,6.5)(8.660254037844386,5.)
\psline[linewidth=0.4pt,linestyle=dashed,dash=2pt 2pt,linecolor=lightgray](8.660254037844386,4.)(6.928203230275509,5.)
\psline[linewidth=0.4pt,linestyle=dashed,dash=2pt 2pt,linecolor=lightgray](6.06217782649107,5.5)(5.196152422706632,6.)
\psline[linewidth=0.4pt,linestyle=dashed,dash=2pt 2pt,linecolor=lightgray](5.196152422706632,6.)(4.330127018922193,6.5)
\psline[linewidth=0.4pt,linestyle=dashed,dash=2pt 2pt,linecolor=lightgray](3.4641016151377544,7.)(2.598076211353316,7.5)
\psline[linewidth=0.4pt,linestyle=dashed,dash=2pt 2pt,linecolor=lightgray](2.598076211353316,6.5)(6.06217782649107,4.5)
\psline[linewidth=0.4pt,linestyle=dashed,dash=2pt 2pt,linecolor=lightgray](6.928203230275509,4.)(7.794228634059947,3.5)
\psline[linewidth=0.4pt,linestyle=dashed,dash=2pt 2pt,linecolor=lightgray](7.794228634059947,3.5)(7.794228634059947,5.5)
\psline[linewidth=0.4pt,linestyle=dashed,dash=2pt 2pt,linecolor=lightgray](2.598076211353316,5.5)(5.196152422706632,4.)
\psline[linewidth=0.4pt,linestyle=dashed,dash=2pt 2pt,linecolor=lightgray](6.06217782649107,3.5)(6.928203230275509,3.)
\psline[linewidth=0.4pt,linestyle=dashed,dash=2pt 2pt,linecolor=lightgray](2.598076211353316,6.5)(6.06217782649107,8.5)
\psline[linewidth=0.4pt,linestyle=dashed,dash=2pt 2pt,linecolor=lightgray](2.598076211353316,5.5)(6.928203230275508,8.)
\psline[linewidth=0.4pt,linestyle=dashed,dash=2pt 2pt,linecolor=lightgray](2.598076211353316,4.5)(4.330127018922193,5.5)
\psline[linewidth=0.4pt,linestyle=dashed,dash=2pt 2pt,linecolor=lightgray](5.196152422706632,6.)(6.06217782649107,6.5)
\psline[linewidth=0.4pt,linestyle=dashed,dash=2pt 2pt,linecolor=lightgray](6.928203230275509,7.)(7.794228634059945,7.5)
\psline[linewidth=0.4pt,linestyle=dashed,dash=2pt 2pt,linecolor=lightgray](8.660254037844386,7.)(7.794228634059947,6.5)
\psline[linewidth=0.4pt,linestyle=dashed,dash=2pt 2pt,linecolor=lightgray](6.928203230275509,6.)(5.196152422706632,5.)
\psline[linewidth=0.4pt,linestyle=dashed,dash=2pt 2pt,linecolor=lightgray](4.330127018922193,4.5)(3.464101615137755,4.)
\psline[linewidth=0.4pt,linestyle=dashed,dash=2pt 2pt,linecolor=lightgray](4.330127018922193,3.5)(8.660254037844386,6.)
\psline[linewidth=0.4pt,linestyle=dashed,dash=2pt 2pt,linecolor=lightgray](8.660254037844386,5.)(5.196152422706632,3.)
\begin{scriptsize}
\rput[bl](-11.684116913639489,10.324547180759712){$Curve$}
\psdots[dotsize=9pt 0,dotstyle=*,linecolor=red](3.4641016151377544,6.)
\psdots[dotsize=9pt 0,dotstyle=*,linecolor=green](3.4641016151377544,7.)
\psdots[dotsize=9pt 0,dotstyle=*,linecolor=red](4.330127018922193,6.5)
\psdots[dotsize=9pt 0,dotstyle=*,linecolor=green](4.330127018922193,7.5)
\psdots[dotsize=9pt 0,dotstyle=*,linecolor=green](5.196152422706632,7.)
\psdots[dotsize=9pt 0,dotstyle=*,linecolor=red](5.196152422706632,6.)
\psdots[dotsize=9pt 0,dotstyle=*,linecolor=red](4.330127018922193,5.5)
\psdots[dotsize=9pt 0,dotstyle=*,linecolor=green](4.330127018922193,4.5)
\psdots[dotsize=9pt 0,dotstyle=*,linecolor=red](5.196152422706632,5.)
\psdots[dotsize=9pt 0,dotstyle=*,linecolor=green](5.196152422706632,4.)
\psdots[dotsize=9pt 0,dotstyle=*,linecolor=green](6.06217782649107,3.5)
\psdots[dotsize=9pt 0,dotstyle=*,linecolor=red](6.06217782649107,4.5)
\psdots[dotsize=9pt 0,dotstyle=*,linecolor=green](6.928203230275509,4.)
\psdots[dotsize=9pt 0,dotstyle=*,linecolor=red](6.928203230275509,5.)
\psdots[dotsize=9pt 0,dotstyle=*,linecolor=red](6.06217782649107,5.5)
\psdots[dotsize=9pt 0,dotstyle=*,linecolor=green](6.06217782649107,6.5)
\psdots[dotsize=9pt 0,dotstyle=*,linecolor=green](6.928203230275509,7.)
\psdots[dotsize=9pt 0,dotstyle=*,linecolor=red](6.928203230275509,6.)
\psdots[dotsize=9pt 0,dotstyle=*,linecolor=green](7.794228634059947,6.5)
\psdots[dotsize=9pt 0,dotstyle=*,linecolor=red](7.794228634059947,5.5)
\end{scriptsize}
\end{pspicture*}
 \caption{ Representation of an feasible conformation for the sequence $(HP)^{2}PH(HP)^{2}(PH)^{2}HP(PH)^{2}$ in the 2D triangular lattice.}
   \label{figure2}
  \end{center}
\end{figure}
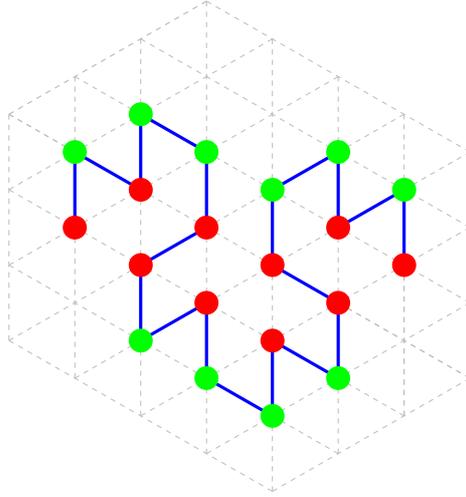

Figure \ref{figure2} represents an feasible  conformation in the 2D triangular lattice model for a protein sequence  of 20 amino acids  given in the HP model by $(HP)^{2}PH(HP)^{2}(PH)^{2}HP(PH)^{2}$. The green points represent the hydrophilic amino acids while the hydrophobic amino acids are represented  in red. The energy of the conformation given in Figure \ref{figure2} is $E(s)=-15$ (15 contacts of H-H type).
 \subsection{Encoding Solution}
 A feasible solution can be represented  by the  sequence of $n-1$ movements in the lattice called self-avoiding paths, (see Figure \ref{figure1})  that generate this conformation. For example, the corresponding movements vector to the solution $S$, given in Figure \ref{figure2} is as follow:
 $$mv(S)=[2,6,2,6,5,4,5,1,5,6,2,6,2,3,2,1,5,1,5].$$
 The sequence $mv(S)$ allows us to deduce the position of each amino acid in the lattice.
\subsection{Calculation of free energy}
Given a feasible protein conformation of $n$ amino acids and let $s$ be its corresponding sequence in the HP model. The folding quality of this of this conformation is measured by the following energy function \cite{lin_efficient_2009}:
\begin{equation*}\label{1}
   E(s)=-\sum_{i=1}^{n-2}\sum_{j=i+2}^{n}\delta_{ij}\ x_{ij},
 \end{equation*}
 where  $\delta_{ij}$ represents an indicator that determines whether both elements $s_{i}$ and $s_{j}$ are hydrophobic.
\begin{equation*}
\delta_{ij}=\left\{
  \begin{array}{ll}
    1; & \hbox{if $(s_{i}=H)\land(s_{j}=H)$}, \vspace{0.1cm}\\
    0; & \hbox{otherwise,}
  \end{array}
\right.
\end{equation*}
and
\begin{equation*}
x_{ij}=\left\{
  \begin{array}{ll}
    1; & \hbox{if the amino acids $i$ and $j$ form an  H-H topological contact,} \vspace{0.1cm}\\
    0; & \hbox{otherwise.}
  \end{array}
\right.
\end{equation*}
\subsection{Mathematical program for the PSP}
In the rest of this section, we present a mathematical program for the Protein Folding Problem (PSP) in its 2D triangular lattice model form. The aim here is to construct a linear mathematical program that can be implemented in modeling languages and compatible with the existing software solvers. Each node in a 2D triangular lattice  blue located in a given position $(i,j)$ has six neighbors represented in a grid on a canonical basis. Considering the following possible neighboring directions of $(i,j)$ node:
\begin{figure}[H]
\begin{center}
\newrgbcolor{ffxfqq}{1. 0.4980392156862745 0.}
\psset{xunit=1.0cm,yunit=1.0cm,algebraic=true,dimen=middle,dotstyle=o,dotsize=5pt 0,linewidth=0.8pt,arrowsize=3pt 2,arrowinset=0.25}
\begin{pspicture*}(0.13739164176913435,1.6934398049439385)(7.671655628542261,6.29710473735538)
\psaxes[labelFontSize=\scriptstyle,xAxis=true,yAxis=true,Dx=1.,Dy=1.,ticksize=-2pt 0,subticks=2]{->}(0,0)(0.13739164176913435,1.6934398049439385)(7.671655628542261,6.29710473735538)
\small{\rput[tl](3.1265760744373963,3.7757853677096027){$(i,j)$}
\rput[tl](2.83,2.683964342577831){$(i+1,j)$}
\rput[tl](4.586168392245131,3.4654783395142568){$(i+1,j+1)$}
\rput[tl](4.609154098037379,4.936563510218118){$(i-1,j+1)$}
\rput[tl](2.83,5.6491203897778){$(i-1,j)$}
\rput[tl](0.45,4.936563510218118){$(i-1,j-1)$}
\rput[tl](0.45,3.4654783395142568){$(i+1,j-1)$}}
\psline[linewidth=0.4pt,linestyle=dashed,dash=1pt 1pt,linecolor=gray](3.4641016151377553,3.)(3.4641016151377544,3.)
\psline[linewidth=0.4pt,linestyle=dashed,dash=1pt 1pt,linecolor=gray](3.4641016151377544,3.)(3.4641016151377544,3.)
\psline[linewidth=0.4pt,linestyle=dashed,dash=1pt 1pt,linecolor=gray](4.330127018922191,3.5)(4.330127018922193,3.5)
\psline[linewidth=0.4pt,linestyle=dashed,dash=1pt 1pt,linecolor=gray](4.330127018922191,4.5)(4.330127018922193,4.5)
\psline[linewidth=0.4pt,linestyle=dashed,dash=1pt 1pt,linecolor=ffxfqq]{->}(3.4641016151377544,4.)(3.464101615137753,5.)
\psline[linewidth=0.4pt,linestyle=dashed,dash=1pt 1pt,linecolor=ffxfqq]{->}(3.4641016151377544,4.)(4.330127018922191,4.5)
\psline[linewidth=0.4pt,linestyle=dashed,dash=1pt 1pt,linecolor=ffxfqq]{->}(3.4641016151377544,4.)(2.5980762113533165,4.5)
\psline[linewidth=0.4pt,linestyle=dashed,dash=1pt 1pt,linecolor=ffxfqq]{->}(3.4641016151377544,4.)(2.5980762113533165,3.5)
\psline[linewidth=0.4pt,linestyle=dashed,dash=1pt 1pt,linecolor=ffxfqq]{->}(3.4641016151377544,4.)(3.4641016151377553,3.)
\psline[linewidth=0.4pt,linestyle=dashed,dash=1pt 1pt,linecolor=ffxfqq]{->}(3.4641016151377544,4.)(4.330127018922191,3.5)
\psline[linewidth=0.4pt,linestyle=dashed,dash=1pt 1pt,linecolor=lightgray](1.7320508075688772,5.)(3.4641016151377544,6.)
\psline[linewidth=0.4pt,linestyle=dashed,dash=1pt 1pt,linecolor=lightgray](3.4641016151377544,6.)(5.196152422706632,5.)
\psline[linewidth=0.4pt,linestyle=dashed,dash=1pt 1pt,linecolor=lightgray](5.196152422706632,5.)(5.196152422706632,3.)
\psline[linewidth=0.4pt,linestyle=dashed,dash=1pt 1pt,linecolor=lightgray](5.196152422706632,3.)(3.4641016151377544,2.)
\psline[linewidth=0.4pt,linestyle=dashed,dash=1pt 1pt,linecolor=lightgray](3.4641016151377544,2.)(1.7320508075688772,3.)
\psline[linewidth=0.4pt,linestyle=dashed,dash=1pt 1pt,linecolor=lightgray](1.7320508075688772,3.)(1.7320508075688772,5.)
\psline[linewidth=0.4pt,linestyle=dashed,dash=1pt 1pt,linecolor=lightgray](1.7320508075688772,4.)(4.330127018922193,5.5)
\psline[linewidth=0.4pt,linestyle=dashed,dash=1pt 1pt,linecolor=lightgray](1.7320508075688772,3.)(2.5980762113533165,3.5)
\psline[linewidth=0.4pt,linestyle=dashed,dash=1pt 1pt,linecolor=lightgray](4.330127018922191,4.5)(5.196152422706632,5.)
\psline[linewidth=0.4pt,linestyle=dashed,dash=1pt 1pt,linecolor=lightgray](2.598076211353316,2.5)(5.196152422706632,4.)
\psline[linewidth=0.4pt,linestyle=dashed,dash=1pt 1pt,linecolor=lightgray](4.330127018922193,2.5)(1.7320508075688772,4.)
\psline[linewidth=0.4pt,linestyle=dashed,dash=1pt 1pt,linecolor=lightgray](5.196152422706632,3.)(4.330127018922191,3.5)
\psline[linewidth=0.4pt,linestyle=dashed,dash=1pt 1pt,linecolor=lightgray](2.5980762113533165,4.5)(1.7320508075688772,5.)
\psline[linewidth=0.4pt,linestyle=dashed,dash=1pt 1pt,linecolor=lightgray](2.598076211353316,5.5)(5.196152422706632,4.)
\psline[linewidth=0.4pt,linestyle=dashed,dash=1pt 1pt,linecolor=lightgray](3.4641016151377544,2.)(3.4641016151377553,3.)
\psline[linewidth=0.4pt,linestyle=dashed,dash=1pt 1pt,linecolor=lightgray](3.4641016151377544,5.)(3.4641016151377544,6.)
\psline[linewidth=0.4pt,linestyle=dashed,dash=1pt 1pt,linecolor=lightgray](2.598076211353316,2.5)(2.598076211353316,5.5)
\psline[linewidth=0.4pt,linestyle=dashed,dash=1pt 1pt,linecolor=lightgray](4.330127018922193,2.5)(4.330127018922193,5.5)
\begin{scriptsize}
\rput[bl](-5.209858125331097,6.986014216552342){$Curve$}
\psdots[dotstyle=*](3.4641016151377544,4.)
\end{scriptsize}
\end{pspicture*}
  \caption{The six positions close to the position $(i,j)$ in the 2D triangular lattice.}
  \label{figure3}
   \end{center}
\end{figure}
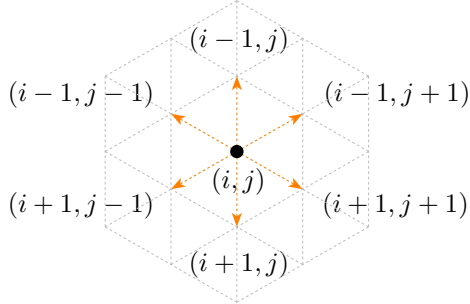
Let $n$ be the length  of the protein sequence, and let $y_{ij}^{k}$ be a three dimensional variable such that:
$$y_{ij}^{k} = \left\{
\begin{array}{ll}
1, & \textrm{if the position }\ (i,j)\ \textrm{contains the}\ k^{th}\ \textrm{amino acid in the protien  sequence},\vspace{0.2cm}\\
0, & \textrm{else}.
\end{array}
\right.
$$
\subsubsection{\textbf{Constraints}}
First, we fix the first amino acid in the protein sequence at position $(n,n)$ as a starting point, i.e., $$y_{nn}^{1}=1.$$
Regarding the problem constraints, we can identify three different constraints that guarantees the admissibility of the resulting solution :
\begin{itemize}
\item[•] A path in the grid is a solution if it occupies exactly $n$ nodes in the grid. This constraint can be written as follows:
$$ \sum_{k=1}^{n}\sum_{i=1}^{2n}\sum_{j=1}^{2n}y_{ij}^{k} = n.$$
\item[•]A node in the grid can contain at most one amino acid  at the $k^{th}$ position, hence:
$$ \sum_{k=1}^{n}y_{ij}^{k} \leq 1,\ \forall i \in \{1,\dots, 2n\},\ \forall j \in \{1,\dots, 2n\}. $$
\item[•]A node in the grid can contain the amino acid  in position $k+1$, if and only if at least one of its neighboring nodes contains the $k^{th}$ amino acid in the protein sequence:\vspace{0.2cm}
$$y_{ij}^{k+1} \leq  y_{i-1j+1}^{k}+y_{i-1j}^{k}+y_{i-1j-1}^{k}+y_{i+1j-1}^{k}+y_{i+1j}^{k}+y_{i+1j+1}^{k},\forall i,j \in \{1,\dots, 2n\},\ \forall k \in \{1,\dots,n-1\}.\vspace{0.2cm}$$
\end{itemize}
\subsubsection{\textbf{The objective function}}
Let $\alpha_{k}$ be a numerical interpretation of any given amino acid  into a binary value (i.e., representation in $\{0,1\}$), where:

$$\alpha_{k} = \left\{
\begin{array}{ll}
1 & \mbox{if the  $k^{th}$ amino acid in the protein  sequence  is hydrophobic, i.e., H,}\vspace{0.2cm}\\
0 & \mbox{if the  $k^{th}$ amino acid in the  protein sequence  is hydrophilic , i.e., P.}
\end{array}
\right.
$$

Thus, the objective function can be calculated  as follows:
$$\max(\mathcal{Z})=\frac{1}{2}\mathcal{Z^{*}}-\dsum_{k=1}^{n-1}\alpha_{k}\alpha_{k+1},$$

where

\begin{small}
$$\hspace{-2mm}\mathcal{Z^{*}}=\max_{y}\left\{\sum_{i=1}^{2n}\sum_{j=1}^{2n}\left(\sum_{k=1}^{n}\alpha_{k}y_{ij}^{k}\right)\left(\sum_{k=1}^{n}\alpha_{k}\left(y_{i-1j+1}^{k}+y_{i-1j}^{k}+y_{i-1j-1}^{k}+y_{i+1j-1}^{k}+y_{i+1j}^{k}+y_{i+1j+1}^{k}\right)\right)\right\}.$$
\end{small}
This mathematical model guaranties optimal solution which is included in the grid enclosed by the points $\{(1,1),(1,2n),(2n,1),(2n,2n)\}$, with a starting point $(n,n)$.
The choice of these limit points is based on the fact that in a path graph $P_n$ on the grid, the maximal distance between the fixed starting node and the rest of the nodes (i.e., basically for the end node) is at most equals to $n-1$  movements. More precisely it represents the radius of the graph, which is the number of edges in the case of a path graph $P_n$. With the view to its spatial complexity $O(n^3)$, this model clearly has  a rather high cost in terms of memory. However, it provides a rather good equilibrium with the computational complexity; since all variables are binary strings/arrays.
\section{Existing 2D H-P protein prediction algorithms }

Recently a number of metaheuristcs have been used to solve the PSP in the 2D triangular lattice model. In \cite{hoque_hybrid_2006}, the authors suggested a new hybrid algorithm, called Hybrid Genetic Algorithm (HGA). This latter enhances the performance of a classical GA implementation by reducing the encountered through the generational process. More specifically, it is clear that when the number of generations increases, the current conformations become very compact. Hence, the application of genetic operators produces invalid conformations with considerable number of collisions. The proposed approach consists in using the generalized short pull moves strategy to produce only valid conformations to avoid blocking GA search processes. The authors have shown considerable quality improvements when compared to a simple genetic algorithm  implementation SGA \cite{hoque_hybrid_2006}.
Later on in \cite{bockenhauer_local_2008}, the authors proposed new approach based on the tabu  search algorithm using  a generalized local move (i.e., pull move) as to improve the landscape exploration and the quality of the produced solutions..
Also, two  approaches are proposed in \cite{lin_efficient_2009}, including the   Elite-based Reproduction Strategy-Genetic Algorithm (ERS-GA) ,and a Hybrid of Hill climbing and Genetic Algorithm called HHGA that combine the ERS-GA with a hill climbing algorithm.A new approach called IMOG has been proposed in \cite{yang_protein_2018}, that  combines ions motion optimization algorithm (IMO) with a Greedy algorithm (G), the obtained results  shown that the IMOG algorithm has a good search ability and stability for the PSP problem using  benchmark data sets.

\section{Novel hybrid approach for  PSP problem}
The most commonly used hybridization in literature consists of combining two metaheuristics, one based on a single solution, known as s-metaheuristics, and the other based on a population, known as p-metaheuristics. The s-metaheuristics have proved their effectiveness for intensification, while the p-metaheuristics known by their exploration capacity. Thus, this type of hybridization allows to establish a balance between the diversification and the intensification of the search process \cite{blum_metaheuristics_2003, raidl_metaheuristic_2010, talbi_metaheuristics:_2009}.

In this work, we proposed a hybrid approach to solve the PSP problem, called GALSTS, in reference to the adopted combination of the following metaheuristics: genetic algorithm, local search, tabu search strategy. As all population-based approaches the first step of GALSTS is to generate an initial population $P$ of $m$ feasible solutions, in which the crossover operator is guided by a tabu list that allows to prohibiting previous movements in order to explore new regions in the search space. Each two selected parent produces a set of offsprings (i.e., $f$ solutions) by applying the crossover procedure controlled by a tabu list that contains all $k $ crossover points that are already used. This has allowed us to avoid cyclical movements.

In order to improve the quality of the children produced by the crossover process, each one of them  is introduced with a probability $p_{m}$ as an initial solution of a local search algorithm. Such that the transition  from a solution $s$ to one of its neighbors $s'$ is carried out by a random choice of an amino acid $i$ and replacing its direction by another one among the five  other possible directions. If the quality of $s$ is better than $s'$, then it becomes the courant solution for the next iteration. This process is repeated until the satisfaction of the  stops criterion.  This improvement phase has allowed us to use the information provided by parents more efficiently to produce high-quality solutions. The two solutions   with the highest fitness value are introduced into the new population $P_{new}$  if they not exist in this population previously, although are of lower quality than their parents to encourage exploration of the research space. The best $m$ solutions of   $P\cup P_{new}$ are replaced the individuals in the population $P$ for the next generation. This approach is intended to avoid the rapid convergence towards local optima, and the wide diversification between the solutions and thus ensures the quality of the solutions during all the stages of the research space explorations, see Algorithm \ref{Algorithm1}.
\begin{algorithm}
\setstretch{1.48}
\caption{Suggested hybrid algorithm GALSTS}
\begin{flushleft}
\algorithmicrequire  A protein sequence of amino acids of size $n$.\vspace{0.1cm} \\
\algorithmicensure  The best confirmation for the protein sequence.\\ \vspace{-0.29cm}
\rule{\linewidth}{0.5pt}
\end{flushleft}
\begin{algorithmic}[H]
\State\hspace{-7mm} \textbf{begin}
 \State  Initialization: $P\leftarrow$ The initial population of $m$ solutions;
 \While {the stop criterion is not checked}
 \State  Create a new population by the flowing steps:
  \State  $k\leftarrow 0,\  P_{new}\leftarrow {\emptyset}$;
  \While {$k \leq m$}
 \State Select two parents $(p_{1},p_{2})$ from $P$;
\State $ offspring \leftarrow$ two solutions  obtained by applying the algorithms (TS-LS) \State \hspace{2.5cm} to the selected parents;
\If {$~ offspring $ does not exist in the new population $P_{new}$}
\State $P_{new}\leftarrow P_{new}\cup \{offspring\}; $
\State $k\leftarrow k+1$;
\EndIf \State{\textbf{end if}}
\EndWhile  \State{\textbf{end while}}
\State $P\leftarrow$  the $m$ best solution among $P\cup P_{new}$;
\EndWhile  \State{\textbf{end while}}
\State\hspace{-7mm}  \textbf{end}\vspace{4mm}
\end{algorithmic}
\label{Algorithm1}
\end{algorithm}
\subsection{The Initial Population}
%We start with an initial population of {\color{blue} $m$ randomly generated individuals}. We put  the first amino acid at one point in the Lattice, {\color{blue}while for the next of amino acid,}  we choose a random direction among the six  directions as shown in Figure \ref{figure1}. If the selected direction generates an already occupied position, we generate another direction different from the selected one. If all the directions create a cycle, we generate a new solution, see Algorithm \ref{Algorithm2}.
We start with an initial population of $m$ randomly generated individuals. An initialization algorithm was proposed that allows to generate only valid conformations for the initial population of GALSTS, we use a list T that contains all the positions used for courant solution,  we put the first amino acid at one point in the lattice and save their position in T. Then, for each amino acid we  choose a random direction among the six  directions as shown in Figure \ref{figure1}. If the selected direction generates an already occupied position (i.e., existing in T); we generate another direction different from the selected one. If all the six directions create an existing positions in T (i.e., all direction create invalid solution), we generate a new solution, see Algorithm \ref{Algorithm2}.
\begin{algorithm}
\setstretch{1.48}
\caption{Generation algorithm for the initial population}
\begin{flushleft}
\algorithmicrequire $n$ the number of amino acids in the protein sequence.\vspace{0.1cm} \\
\algorithmicensure A feasible confirmation for the protein sequence.\\ \vspace{-0.29cm}
\rule{\linewidth}{0.5pt}
\end{flushleft}
\begin{algorithmic}
\State \hspace{-7mm}  \textbf{begin}
 \State Initialization: $T\leftarrow$ Table of position for each amino acids in lattice;
 \State  $T[1]\leftarrow (x_{1},y_{1}),$ i.e., put the first monomer in one vertex  of the; lattice
 \State $i\leftarrow 2;$
 \While {($i \leq n $)}
 \State $K\leftarrow \{1,2,3,4,6\};$ the set of all possible directions on the 2D triangular lattice
 \State  $t\leftarrow true; $
\While {($t = true)$}
\State $u\leftarrow$ Random direction generated from the list  $K$;
\State $K\leftarrow K\backslash\{ u \}$;
\State $(x_{i},y_{i})\leftarrow$ Position generated by the direction $u$;
\If{ $(x_{i},y_{i})\notin T$}
\State $t \leftarrow false$;
\State $T[i]\leftarrow (x_{i},y_{i})$;
\ElsIf{($K=\emptyset$)}
\State $ t\leftarrow false$; \State $i\leftarrow 2$;
\EndIf  \State{\textbf{end if}}
\EndWhile  \State{\textbf{end while}}
\EndWhile  \State{\textbf{end while}}
\State\hspace{-7mm}  \textbf{end}\vspace{4mm}
\end{algorithmic}
\label{Algorithm2}
\end{algorithm}

 \subsection{ Crossover Operator}
\begin{figure}
\begin{center}
\psset{xunit=1.0cm,yunit=1.0cm,algebraic=true,dimen=middle,dotstyle=o,dotsize=5pt 0,linewidth=0.8pt,arrowsize=3pt 2,arrowinset=0.25}
\begin{pspicture*}(2.0601631052838894,0.9360361213431786)(15.361313098544377,13.436136465149408)
\psline[linewidth=1.2pt,linecolor=blue](11.258330249197702,5.5)(12.12435565298214,6.)
\psline[linewidth=1.2pt,linecolor=blue](12.12435565298214,6.)(11.248430027361016,6.50463699695503)
\psline[linewidth=1.2pt,linecolor=blue](12.990381056766578,4.5)(12.12435565298214,4.)
\psline[linewidth=1.2pt,linecolor=blue](12.12435565298214,4.)(12.990381056766578,3.5)
\psline[linewidth=1.2pt,linecolor=blue](12.990381056766578,3.5)(12.12435565298214,3.)
\psline[linewidth=1.2pt,linecolor=blue](12.12435565298214,3.)(11.258330249197702,3.5)
\psline[linewidth=1.2pt,linecolor=blue](11.258330249197702,3.5)(10.392304845413264,3.)
\psline{->}(8.660254037844386,8.)(8.660254037844386,7.)
\psline[linewidth=1.2pt,linecolor=blue](10.392304845413264,3.)(10.392304845413264,4.)
\psline[linewidth=1.2pt,linecolor=blue](10.392304845413264,9.)(11.258330249197702,9.5)
\psline[linewidth=1.2pt,linecolor=blue](11.258330249197702,9.5)(11.258330249197702,8.5)
\psline[linewidth=1.2pt,linecolor=blue](11.258330249197702,8.5)(12.12435565298214,9.)
\psline[linewidth=1.2pt,linecolor=blue](12.12435565298214,9.)(12.990381056766578,8.5)
\psline[linewidth=1.2pt,linecolor=blue](12.990381056766578,8.5)(13.856406460551018,9.)
\psline[linewidth=1.2pt,linecolor=blue](13.856406460551018,9.)(11.258330249197702,10.5)
\psline[linewidth=1.2pt,linecolor=blue](11.258330249197702,10.5)(12.12435565298214,11.)
\psline[linewidth=1.2pt,linecolor=blue](12.12435565298214,11.)(11.25186339099227,11.50090005189002)
\psline[linewidth=1.2pt,linecolor=blue](3.4589282786474254,9.994898896197055)(5.196152422706632,11.)
\psline[linewidth=1.2pt,linecolor=blue](5.196152422706632,11.)(6.928203230275509,10.)
\psline[linewidth=1.2pt,linecolor=blue](6.928203230275509,10.)(6.06217782649107,9.5)
\psline[linewidth=1.2pt,linecolor=blue](6.06217782649107,9.5)(6.928203230275509,9.)
\psline[linewidth=1.2pt,linecolor=blue](6.928203230275509,9.)(6.06217782649107,8.5)
\psline[linewidth=1.2pt,linecolor=blue](6.06217782649107,8.5)(5.196152422706632,9.)
\psline[linewidth=1.2pt,linecolor=blue](5.196152422706632,9.)(4.330127018922193,8.5)
\psline[linewidth=1.2pt,linecolor=blue](4.330127018922193,8.5)(4.330127018922193,9.5)
\psline[linewidth=1.2pt,linecolor=blue](3.459852276881338,4.997218055534968)(5.1919030844502165,5.997218055534968)
\psline[linewidth=1.2pt,linecolor=blue](5.1919030844502165,5.997218055534968)(6.923953892019093,4.997218055534968)
\psline[linewidth=1.2pt,linecolor=blue](6.923953892019093,4.997218055534968)(7.794228634059947,4.5)
\psline[linewidth=1.2pt,linecolor=blue](7.794228634059947,4.5)(6.928203230275509,4.)
\psline[linewidth=1.2pt,linecolor=blue](6.928203230275509,4.)(6.06217782649107,4.5)
\psline[linewidth=1.2pt,linecolor=blue](6.06217782649107,4.5)(5.196152422706632,4.)
\psline[linewidth=1.2pt,linecolor=blue](5.196152422706632,4.)(5.196152422706632,5.)
\psline[linewidth=1.2pt,linecolor=blue](5.196152422706632,5.)(4.330127018922193,4.5)
\psline[linewidth=1.2pt,linecolor=blue](12.990381056766578,4.5)(11.258330249197702,5.5)
\psline[linewidth=1.2pt](8.660254037844386,8.)(8.660254037844386,7.182989554445888)
\pscircle(13.027527128527113,9.478553705469718){0.2790444719430972}
\pscircle(6.928203230275509,10.){0.2854705966425739}
\rput[tl](5.052454423685197,12.26){$p_{1}$}
\rput[tl](11.050615260832204,12.26){$p_{2}$}
\rput[tl](11,7.17){$f_{2}$}
\rput[tl](5,7.17){$f_{1}$}
\psline[linewidth=0.4pt,linestyle=dashed,dash=3pt 3pt,linecolor=lightgray](2.598076211353316,11.5)(5.196152422706632,13.)
\psline[linewidth=0.4pt,linestyle=dashed,dash=3pt 3pt,linecolor=lightgray](2.598076211353316,11.5)(2.598076211353316,2.5)
\psline[linewidth=0.4pt,linestyle=dashed,dash=3pt 3pt,linecolor=lightgray](2.598076211353316,2.5)(4.330127018922193,1.5)
\psline[linewidth=0.4pt,linestyle=dashed,dash=3pt 3pt,linecolor=lightgray](2.598076211353316,3.5)(6.06217782649107,1.5)
\psline[linewidth=0.4pt,linestyle=dashed,dash=3pt 3pt,linecolor=lightgray](2.598076211353316,4.5)(7.794228634059947,1.5)
\psline[linewidth=0.4pt,linestyle=dashed,dash=3pt 3pt,linecolor=lightgray](2.598076211353316,5.5)(9.526279441628825,1.5)
\psline[linewidth=0.4pt,linestyle=dashed,dash=3pt 3pt,linecolor=lightgray](2.598076211353316,6.5)(6.06217782649107,4.5)
\psline[linewidth=0.4pt,linestyle=dashed,dash=3pt 3pt,linecolor=lightgray](6.928203230275509,4.)(11.258330249197702,1.5)
\psline[linewidth=0.4pt,linestyle=dashed,dash=3pt 3pt,linecolor=lightgray](2.598076211353316,7.5)(5.1919030844502165,5.997218055534968)
\psline[linewidth=0.4pt,linestyle=dashed,dash=3pt 3pt,linecolor=lightgray](7.794228634059947,4.5)(12.990381056766578,1.5)
\psline[linewidth=0.4pt,linestyle=dashed,dash=3pt 3pt,linecolor=lightgray](2.598076211353316,8.5)(11.258330249197702,3.5)
\psline[linewidth=0.4pt,linestyle=dashed,dash=3pt 3pt,linecolor=lightgray](2.598076211353316,9.5)(12.12435565298214,4.)
\psline[linewidth=0.4pt,linestyle=dashed,dash=3pt 3pt,linecolor=lightgray](12.990381056766578,3.5)(14.722431864335457,2.5)
\psline[linewidth=0.4pt,linestyle=dashed,dash=3pt 3pt,linecolor=lightgray](12.990381056766578,4.5)(14.722431864335457,3.5)
\psline[linewidth=0.4pt,linestyle=dashed,dash=3pt 3pt,linecolor=lightgray](11.258330249197702,5.5)(6.06217782649107,8.5)
\psline[linewidth=0.4pt,linestyle=dashed,dash=3pt 3pt,linecolor=lightgray](5.196152422706632,9.)(2.598076211353316,10.5)
\psline[linewidth=0.4pt,linestyle=dashed,dash=3pt 3pt,linecolor=lightgray](2.598076211353316,11.5)(6.06217782649107,9.5)
\psline[linewidth=0.4pt,linestyle=dashed,dash=3pt 3pt,linecolor=lightgray](6.928203230275509,9.)(11.248430027361016,6.50463699695503)
\psline[linewidth=0.4pt,linestyle=dashed,dash=3pt 3pt,linecolor=lightgray](12.12435565298214,6.)(14.722431864335457,4.5)
\psline[linewidth=0.4pt,linestyle=dashed,dash=3pt 3pt,linecolor=lightgray](14.722431864335457,5.5)(6.928203230275509,10.)
\psline[linewidth=0.4pt,linestyle=dashed,dash=3pt 3pt,linecolor=lightgray](5.196152422706632,11.)(3.4641016151377544,12.)
\psline[linewidth=0.4pt,linestyle=dashed,dash=3pt 3pt,linecolor=lightgray](4.330127018922193,12.5)(14.722431864335457,6.5)
\psline[linewidth=0.4pt,linestyle=dashed,dash=3pt 3pt,linecolor=lightgray](14.722431864335457,7.5)(12.990381056766578,8.5)
\psline[linewidth=0.4pt,linestyle=dashed,dash=3pt 3pt,linecolor=lightgray](11.258330249197702,9.5)(5.196152422706632,13.)
\psline[linewidth=0.4pt,linestyle=dashed,dash=3pt 3pt,linecolor=lightgray](6.928203230275509,13.)(11.258330249197702,10.5)
\psline[linewidth=0.4pt,linestyle=dashed,dash=3pt 3pt,linecolor=lightgray](13.856406460551018,9.)(14.722431864335457,8.5)
\psline[linewidth=0.4pt,linestyle=dashed,dash=3pt 3pt,linecolor=lightgray](14.722431864335457,9.5)(12.12435565298214,11.)
\psline[linewidth=0.4pt,linestyle=dashed,dash=3pt 3pt,linecolor=lightgray](11.25186339099227,11.50090005189002)(8.660254037844386,13.)
\psline[linewidth=0.4pt,linestyle=dashed,dash=3pt 3pt,linecolor=lightgray](10.392304845413264,13.)(14.722431864335457,10.5)
\psline[linewidth=0.4pt,linestyle=dashed,dash=3pt 3pt,linecolor=lightgray](14.722431864335457,11.5)(12.12435565298214,13.)
\psline[linewidth=0.4pt,linestyle=dashed,dash=3pt 3pt,linecolor=lightgray](6.928203230275509,13.)(2.598076211353316,10.5)
\psline[linewidth=0.4pt,linestyle=dashed,dash=3pt 3pt,linecolor=lightgray](8.660254037844386,13.)(5.196152422706632,11.)
\psline[linewidth=0.4pt,linestyle=dashed,dash=3pt 3pt,linecolor=lightgray](3.4589282786474254,9.994898896197055)(2.598076211353316,9.5)
\psline[linewidth=0.4pt,linestyle=dashed,dash=3pt 3pt,linecolor=lightgray](2.598076211353316,8.5)(10.392304845413264,13.)
\psline[linewidth=0.4pt,linestyle=dashed,dash=3pt 3pt,linecolor=lightgray](12.12435565298214,13.)(6.928203230275509,10.)
\psline[linewidth=0.4pt,linestyle=dashed,dash=3pt 3pt,linecolor=lightgray](6.06217782649107,9.5)(5.196152422706632,9.)
\psline[linewidth=0.4pt,linestyle=dashed,dash=3pt 3pt,linecolor=lightgray](4.330127018922193,8.5)(2.598076211353316,7.5)
\psline[linewidth=0.4pt,linestyle=dashed,dash=3pt 3pt,linecolor=lightgray](2.598076211353316,6.5)(6.06217782649107,8.5)
\psline[linewidth=0.4pt,linestyle=dashed,dash=3pt 3pt,linecolor=lightgray](6.928203230275509,9.)(12.990381056766578,12.5)
\psline[linewidth=0.4pt,linestyle=dashed,dash=3pt 3pt,linecolor=lightgray](13.856406460551018,12.)(12.12435565298214,11.)
\psline[linewidth=0.4pt,linestyle=dashed,dash=3pt 3pt,linecolor=lightgray](11.258330249197702,10.5)(2.598076211353316,5.5)
\psline[linewidth=0.4pt,linestyle=dashed,dash=3pt 3pt,linecolor=lightgray](2.598076211353316,4.5)(3.459852276881338,4.997218055534968)
\psline[linewidth=0.4pt,linestyle=dashed,dash=3pt 3pt,linecolor=lightgray](5.1919030844502165,5.997218055534968)(10.392304845413264,9.)
\psline[linewidth=0.4pt,linestyle=dashed,dash=3pt 3pt,linecolor=lightgray](11.258330249197702,9.5)(14.722431864335457,11.5)
\psline[linewidth=0.4pt,linestyle=dashed,dash=3pt 3pt,linecolor=lightgray](14.722431864335457,10.5)(12.12435565298214,9.)
\psline[linewidth=0.4pt,linestyle=dashed,dash=3pt 3pt,linecolor=lightgray](11.258330249197702,8.5)(5.196152422706632,5.)
\psline[linewidth=0.4pt,linestyle=dashed,dash=3pt 3pt,linecolor=lightgray](4.330127018922193,4.5)(2.598076211353316,3.5)
\psline[linewidth=0.4pt,linestyle=dashed,dash=3pt 3pt,linecolor=lightgray](2.598076211353316,2.5)(5.196152422706632,4.)
\psline[linewidth=0.4pt,linestyle=dashed,dash=3pt 3pt,linecolor=lightgray](6.06217782649107,4.5)(12.990381056766578,8.5)
\psline[linewidth=0.4pt,linestyle=dashed,dash=3pt 3pt,linecolor=lightgray](13.856406460551018,9.)(14.722431864335457,9.5)
\psline[linewidth=0.4pt,linestyle=dashed,dash=3pt 3pt,linecolor=lightgray](14.722431864335457,8.5)(7.794228634059947,4.5)
\psline[linewidth=0.4pt,linestyle=dashed,dash=3pt 3pt,linecolor=lightgray](14.722431864335457,7.5)(12.12435565298214,6.)
\psline[linewidth=0.4pt,linestyle=dashed,dash=3pt 3pt,linecolor=lightgray](11.258330249197702,5.5)(4.330127018922193,1.5)
\psline[linewidth=0.4pt,linestyle=dashed,dash=3pt 3pt,linecolor=lightgray](3.464101615137754,2.)(6.928203230275509,4.)
\psline[linewidth=0.4pt,linestyle=dashed,dash=3pt 3pt,linecolor=lightgray](3.4641016151377544,12.)(3.464101615137754,2.)
\psline[linewidth=0.4pt,linestyle=dashed,dash=3pt 3pt,linecolor=lightgray](4.330127018922193,12.5)(4.330127018922193,9.5)
\psline[linewidth=0.4pt,linestyle=dashed,dash=3pt 3pt,linecolor=lightgray](4.330127018922193,8.5)(4.330127018922193,1.5)
\psline[linewidth=0.4pt,linestyle=dashed,dash=3pt 3pt,linecolor=lightgray](5.196152422706632,13.)(5.196152422706632,5.)
\psline[linewidth=0.4pt,linestyle=dashed,dash=3pt 3pt,linecolor=lightgray](5.196152422706632,4.)(5.196152422706631,2.)
\psline[linewidth=0.4pt,linestyle=dashed,dash=3pt 3pt,linecolor=lightgray](6.06217782649107,1.5)(6.062177826491068,12.5)
\psline[linewidth=0.4pt,linestyle=dashed,dash=3pt 3pt,linecolor=lightgray](6.928203230275509,13.)(6.928203230275508,2.)
\psline[linewidth=0.4pt,linestyle=dashed,dash=3pt 3pt,linecolor=lightgray](7.794228634059947,1.5)(7.794228634059944,12.5)
\psline[linewidth=0.4pt,linestyle=dashed,dash=3pt 3pt,linecolor=lightgray](8.660254037844386,13.)(8.660254037844386,7.999567451524234)
\psline[linewidth=0.4pt,linestyle=dashed,dash=3pt 3pt,linecolor=lightgray](8.660254037844387,7.)(8.660254037844386,2.)
\psline[linewidth=0.4pt,linestyle=dashed,dash=3pt 3pt,linecolor=lightgray](9.526279441628825,1.5)(9.525460189139906,12.499527004354988)
\psline[linewidth=0.4pt,linestyle=dashed,dash=3pt 3pt,linecolor=lightgray](10.392304845413264,13.)(10.392304845413264,4.)
\psline[linewidth=0.4pt,linestyle=dashed,dash=3pt 3pt,linecolor=lightgray](10.392304845413264,3.)(10.392304845413264,2.)
\psline[linewidth=0.4pt,linestyle=dashed,dash=3pt 3pt,linecolor=lightgray](11.258330249197702,1.5)(11.258330249197702,8.5)
\psline[linewidth=0.4pt,linestyle=dashed,dash=3pt 3pt,linecolor=lightgray](11.258330249197702,9.5)(11.258330249197702,12.5)
\psline[linewidth=0.4pt,linestyle=dashed,dash=3pt 3pt,linecolor=lightgray](12.12435565298214,13.)(12.12435565298214,2.)
\psline[linewidth=0.4pt,linestyle=dashed,dash=3pt 3pt,linecolor=lightgray](12.990381056766578,1.5)(12.990381056766578,12.5)
\psline[linewidth=0.4pt,linestyle=dashed,dash=3pt 3pt,linecolor=lightgray](13.856406460551018,12.)(13.856406460551018,2.)
\psline[linewidth=0.4pt,linestyle=dashed,dash=3pt 3pt,linecolor=lightgray](14.722431864335457,11.5)(14.722431864335457,2.5)
\psline[linewidth=0.4pt,linestyle=dashed,dash=3pt 3pt,linecolor=lightgray](12.990381056766578,1.5)(14.722431864335457,2.5)
\psline[linewidth=0.4pt,linestyle=dashed,dash=3pt 3pt,linecolor=lightgray](12.12435565298214,3.)(13.8361441601704,1.9883015554208523)
\psline[linewidth=0.4pt,linestyle=dashed,dash=3pt 3pt,linecolor=lightgray](11.258330249197702,1.5)(14.722431864335457,3.5)
\psline[linewidth=0.4pt,linestyle=dashed,dash=3pt 3pt,linecolor=lightgray](9.526279441628823,1.5)(12.12435565298214,3.)
\psline[linewidth=0.4pt,linestyle=dashed,dash=3pt 3pt,linecolor=lightgray](12.990381056766578,3.5)(14.722431864335457,4.452439456690773)
\psline[linewidth=0.4pt,linestyle=dashed,dash=3pt 3pt,linecolor=lightgray](7.794228634059947,1.5)(10.392304845413264,3.)
\psline[linewidth=0.4pt,linestyle=dashed,dash=3pt 3pt,linecolor=lightgray](6.0621778264910695,1.5)(14.722431864335457,6.5)
\psline[linewidth=0.4pt,linestyle=dashed,dash=3pt 3pt,linecolor=lightgray](11.258330249197702,3.5)(12.12435565298214,4.)
\psline[linewidth=0.4pt,linestyle=dashed,dash=3pt 3pt,linecolor=lightgray](12.990381056766578,4.5)(14.722431864335457,5.5)
\begin{scriptsize}
\psdots[dotsize=7pt 0,dotstyle=*,linecolor=green](11.258330249197702,5.5)
\psdots[dotsize=7pt 0,dotstyle=*,linecolor=red](12.12435565298214,6.)
\psdots[dotsize=7pt 0,dotstyle=*,linecolor=red](11.248430027361016,6.50463699695503)
\psdots[dotsize=7pt 0,dotstyle=*,linecolor=red](12.990381056766578,4.5)
\psdots[dotsize=7pt 0,dotstyle=*,linecolor=green](12.12435565298214,4.)
\psdots[dotsize=7pt 0,dotstyle=*,linecolor=green](12.990381056766578,3.5)
\psdots[dotsize=7pt 0,dotstyle=*,linecolor=red](12.12435565298214,3.)
\psdots[dotsize=7pt 0,dotstyle=*,linecolor=green](11.258330249197702,3.5)
\psdots[dotsize=7pt 0,dotstyle=*,linecolor=red](10.392304845413264,3.)
\psdots[dotsize=7pt 0,dotstyle=*,linecolor=red](10.392304845413264,4.)
\psdots[dotsize=7pt 0,dotstyle=*,linecolor=red](10.392304845413264,9.)
\psdots[dotsize=7pt 0,dotstyle=*,linecolor=red](11.258330249197702,9.5)
\psdots[dotsize=7pt 0,dotstyle=*,linecolor=green](11.258330249197702,8.5)
\psdots[dotsize=7pt 0,dotstyle=*,linecolor=red](12.12435565298214,9.)
\psdots[dotsize=7pt 0,dotstyle=*,linecolor=green](12.990381056766578,8.5)
\psdots[dotsize=7pt 0,dotstyle=*,linecolor=green](13.856406460551018,9.)
\psdots[dotsize=7pt 0,dotstyle=*,linecolor=green](11.258330249197702,10.5)
\psdots[dotsize=7pt 0,dotstyle=*,linecolor=red](12.12435565298214,11.)
\psdots[dotsize=7pt 0,dotstyle=*,linecolor=red](11.25186339099227,11.50090005189002)
\psdots[dotsize=7pt 0,dotstyle=*,linecolor=red](12.161501724742694,9.978553705469707)
\psdots[dotsize=7pt 0,dotstyle=*,linecolor=red](13.027527128527113,9.478553705469718)
\psdots[dotsize=7pt 0,dotstyle=*,linecolor=red](3.4589282786474254,9.994898896197055)
\psdots[dotsize=7pt 0,dotstyle=*,linecolor=green](5.196152422706632,11.)
\psdots[dotsize=7pt 0,dotstyle=*,linecolor=red](6.928203230275509,10.)
\psdots[dotsize=7pt 0,dotstyle=*,linecolor=green](6.06217782649107,9.5)
\psdots[dotsize=7pt 0,dotstyle=*,linecolor=green](6.928203230275509,9.)
\psdots[dotsize=7pt 0,dotstyle=*,linecolor=red](6.06217782649107,8.5)
\psdots[dotsize=7pt 0,dotstyle=*,linecolor=green](5.196152422706632,9.)
\psdots[dotsize=7pt 0,dotstyle=*,linecolor=red](4.330127018922193,8.5)
\psdots[dotsize=7pt 0,dotstyle=*,linecolor=red](4.330127018922193,9.5)
\psdots[dotsize=7pt 0,dotstyle=*,linecolor=red](4.327540350677035,10.497449448098532)
\psdots[dotsize=7pt 0,dotstyle=*,linecolor=red](6.062177826491076,10.5)
\psdots[dotsize=7pt 0,dotstyle=*,linecolor=red](3.459852276881338,4.997218055534968)
\psdots[dotsize=7pt 0,dotstyle=*,linecolor=green](5.1919030844502165,5.997218055534968)
\psdots[dotsize=7pt 0,dotstyle=*,linecolor=red](6.923953892019093,4.997218055534968)
\psdots[dotsize=7pt 0,dotstyle=*,linecolor=red](4.375028315760396,5.525595187937687)
\psdots[dotsize=7pt 0,dotstyle=*,linecolor=red](6.063777853140002,5.493840923132277)
\psdots[dotsize=7pt 0,dotstyle=*,linecolor=green](7.794228634059947,4.5)
\psdots[dotsize=7pt 0,dotstyle=*,linecolor=green](6.928203230275509,4.)
\psdots[dotsize=7pt 0,dotstyle=*,linecolor=red](6.06217782649107,4.5)
\psdots[dotsize=7pt 0,dotstyle=*,linecolor=green](5.196152422706632,4.)
\psdots[dotsize=7pt 0,dotstyle=*,linecolor=red](5.196152422706632,5.)
\psdots[dotsize=7pt 0,dotstyle=*,linecolor=red](4.330127018922193,4.5)
\psdots[dotsize=7pt 0,dotstyle=*,linecolor=red](12.124355652982134,5.)
\end{scriptsize}
\end{pspicture*}
  \caption{Crossover operator applied on two conformations of the sequence $H^{2}PH^{2}P^{2}HPH^{2}$. The circled nodes indicate the cutting points positions.}
  \label{figure4}
   \end{center}
\end{figure}
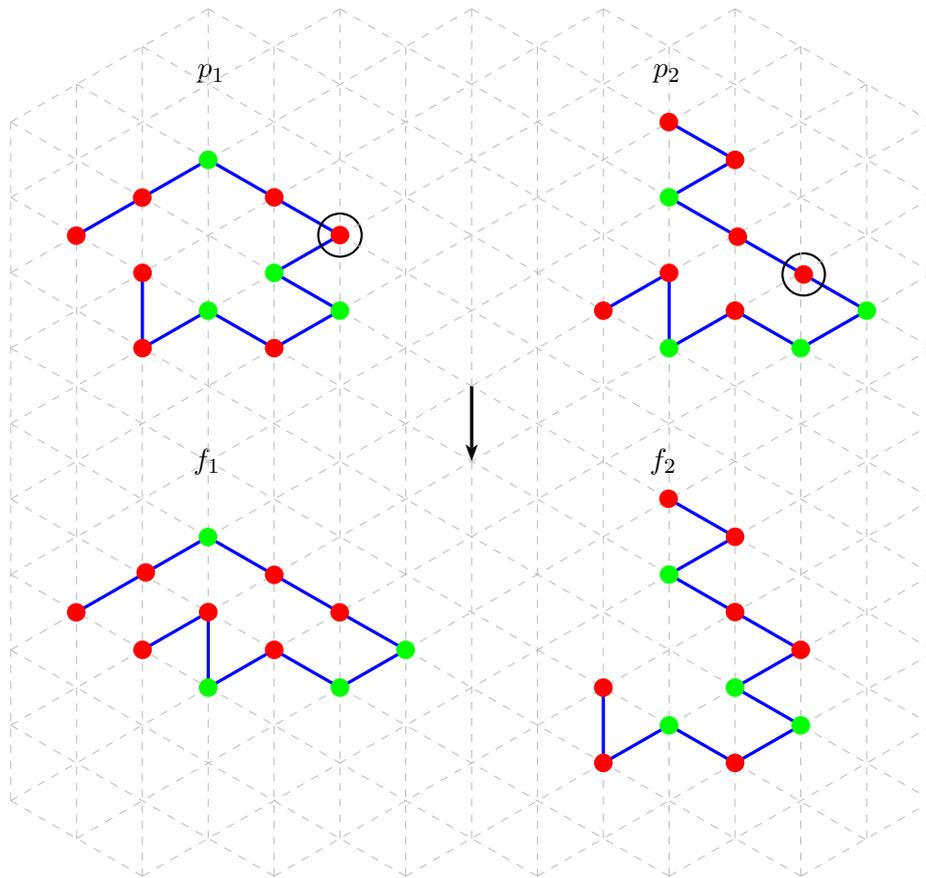

The crossover procedure consists to combine two or more solutions, called parents, as to create other solutions, called offsprings. There are several types of the crossover operator, here we opted for a random 1-point, which consists to swap after selecting two parents $p_{1}$ and $p_{2}$, and generating one random point $c_{1}$, $1 < c_{1} < n$, the parent subsequences limited by $c_{1}$ and $n$. As shown in Figure \ref{figure4}, the new two  conformations (offsprings) $f_{1}$ and $f_{2}$, are obtained by combining $p_{1}$ and $p_{2}$ after generating a random cut point (here $c_{1}=5$). The energy value of offspring $f_{1}$ is $E(f_{1})=-7$, lower than the energy values of its parents, $E(p_{1})=-2$ and $E(p_{2})=-5$.
\subsection{ Mutation Operator}
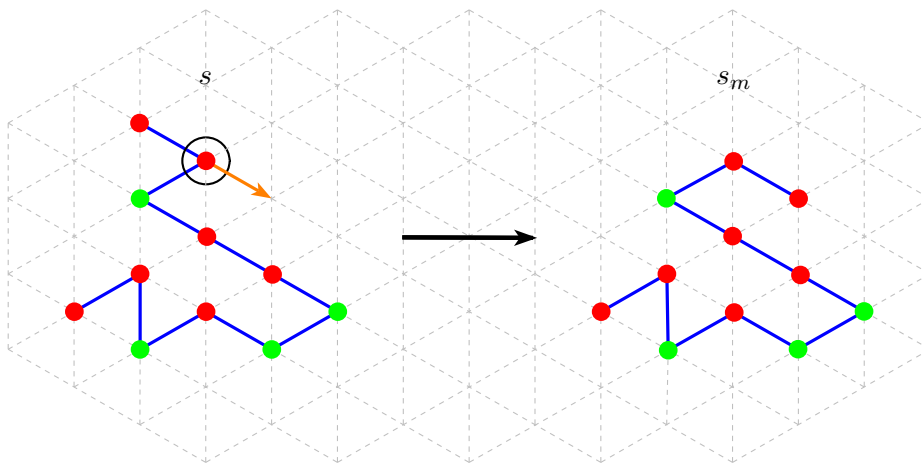
\begin{figure}[H]
\newrgbcolor{ffxfqq}{1. 0.4980392156862745 0.}
\psset{xunit=1.0cm,yunit=1.0cm,algebraic=true,dimen=middle,dotstyle=o,dotsize=5pt 0,linewidth=0.8pt,arrowsize=3pt 2,arrowinset=0.25}
\begin{pspicture*}(2.4337661994090416,5.857344955348472)(14.866931476362668,12.183216016631928)
\psaxes[labelFontSize=\scriptstyle,xAxis=true,yAxis=true,Dx=1.,Dy=1.,ticksize=-2pt 0,subticks=2]{->}(0,0)(2.4337661994090416,5.857344955348472)(14.866931476362668,12.183216016631928)
\psline[linewidth=1.2pt,linecolor=blue](3.4670448157042264,7.998688347875905)(4.333070219488664,8.498688347875907)
\psline[linewidth=1.2pt,linecolor=blue](4.333070219488664,8.498688347875907)(4.333070219488664,7.498688347875906)
\psline[linewidth=1.2pt,linecolor=blue](4.333070219488664,7.498688347875906)(5.199095623273103,7.998688347875905)
\psline[linewidth=1.2pt,linecolor=blue](5.199095623273103,7.998688347875905)(6.065121027057539,7.498688347875906)
\psline[linewidth=1.2pt,linecolor=blue](6.065121027057539,7.498688347875906)(6.931146430841982,7.998688347875905)
\psline[linewidth=1.2pt,linecolor=blue](6.931146430841982,7.998688347875905)(4.333070219488664,9.498688347875905)
\psline[linewidth=1.2pt,linecolor=blue](4.333070219488664,9.498688347875905)(5.199095623273103,9.998688347875905)
\psline[linewidth=1.2pt,linecolor=blue](5.199095623273103,9.998688347875905)(4.326603361283232,10.499588399765926)
\pscircle(5.199095623273103,9.998688347875905){0.31074756589840613}
\psline[linewidth=1.2pt,,linecolor=ffxfqq]{->}(5.199095623273103,9.998688347875905)(6.0621778264910695,9.5)
\psline[linewidth=1.2pt,linecolor=blue](10.395370205782715,7.998230213365534)(11.260782537493263,8.498584170692427)
\psline[linewidth=1.2pt,linecolor=blue](11.260782537493263,8.498584170692427)(11.277789043465686,7.488765459891273)
\psline[linewidth=1.2pt,linecolor=blue](11.277789043465686,7.488765459891273)(12.147645281899102,7.9865537264754645)
\psline[linewidth=1.2pt,linecolor=blue](12.147645281899102,7.9865537264754645)(12.985899331559215,7.502587525254903)
\psline[linewidth=1.2pt,linecolor=blue](12.985899331559215,7.502587525254903)(13.854165597947336,8.00129376262745)
\psline[linewidth=1.2pt,linecolor=blue](13.854165597947336,8.00129376262745)(13.01750152033252,8.484341993059656)
\psline[linewidth=1.2pt,linecolor=blue](13.01750152033252,8.484341993059656)(12.126194869203811,8.99893812801932)
\psline[linewidth=1.2pt,linecolor=blue](12.126194869203811,8.99893812801932)(11.254750528026433,9.502066752981854)
\psline[linewidth=1.2pt,linecolor=blue](11.254750528026433,9.502066752981854)(12.138479224601872,9.99184575212343)
\psline[linewidth=1.2pt,linecolor=blue](12.138479224601872,9.99184575212343)(12.99160720091437,9.49929208534622)
\psline[linewidth=0.4pt,linestyle=dashed,dash=2pt 2pt,linecolor=lightgray](2.598076211353316,10.5)(5.196152422706632,12.)
\psline[linewidth=0.4pt,linestyle=dashed,dash=2pt 2pt,linecolor=lightgray](2.598076211353316,9.5)(6.928203230275509,12.)
\psline[linewidth=0.4pt,linestyle=dashed,dash=2pt 2pt,linecolor=lightgray](2.598076211353316,8.5)(4.333070219488664,9.498688347875905)
\psline[linewidth=0.4pt,linestyle=dashed,dash=2pt 2pt,linecolor=lightgray](5.199095623273103,9.998688347875905)(8.660254037844386,12.)
\psline[linewidth=0.4pt,linestyle=dashed,dash=2pt 2pt,linecolor=lightgray](2.598076211353316,7.5)(3.4670448157042264,7.998688347875905)
\psline[linewidth=0.4pt,linestyle=dashed,dash=2pt 2pt,linecolor=lightgray](4.333070219488664,8.498688347875907)(10.392304845413264,12.)
\psline[linewidth=0.4pt,linestyle=dashed,dash=2pt 2pt,linecolor=lightgray](3.4641016151377544,7.)(4.333070219488664,7.498688347875906)
\psline[linewidth=0.4pt,linestyle=dashed,dash=2pt 2pt,linecolor=lightgray](5.199095623273103,7.998688347875905)(12.12435565298214,12.)
\psline[linewidth=0.4pt,linestyle=dashed,dash=2pt 2pt,linecolor=lightgray](4.330127018922193,6.5)(6.065121027057539,7.498688347875906)
\psline[linewidth=0.4pt,linestyle=dashed,dash=2pt 2pt,linecolor=lightgray](6.931146430841982,7.998688347875905)(12.990381056766578,11.5)
\psline[linewidth=0.4pt,linestyle=dashed,dash=2pt 2pt,linecolor=lightgray](5.196152422706632,6.)(11.254750528026433,9.502066752981854)
\psline[linewidth=0.4pt,linestyle=dashed,dash=2pt 2pt,linecolor=lightgray](12.138479224601872,9.99184575212343)(13.856406460551018,11.)
\psline[linewidth=0.4pt,linestyle=dashed,dash=2pt 2pt,linecolor=lightgray](6.928203230275509,6.)(10.395370205782715,7.998230213365534)
\psline[linewidth=0.4pt,linestyle=dashed,dash=2pt 2pt,linecolor=lightgray](11.260782537493263,8.498584170692427)(14.722431864335457,10.5)
\psline[linewidth=0.4pt,linestyle=dashed,dash=2pt 2pt,linecolor=lightgray](8.660254037844386,6.)(11.277789043465686,7.488765459891273)
\psline[linewidth=0.4pt,linestyle=dashed,dash=2pt 2pt,linecolor=lightgray](12.147645281899102,7.9865537264754645)(14.722431864335457,9.5)
\psline[linewidth=0.4pt,linestyle=dashed,dash=2pt 2pt,linecolor=lightgray](10.392304845413264,6.)(12.985899331559215,7.502587525254903)
\psline[linewidth=0.4pt,linestyle=dashed,dash=2pt 2pt,linecolor=lightgray](13.854165597947336,8.00129376262745)(14.722431864335457,8.5)
\psline[linewidth=0.4pt,linestyle=dashed,dash=2pt 2pt,linecolor=lightgray](12.12435565298214,6.)(14.722431864335457,7.5)
\psline[linewidth=0.4pt,linestyle=dashed,dash=2pt 2pt,linecolor=lightgray](10.392304845413264,12.)(14.722431864335457,9.5)
\psline[linewidth=0.4pt,linestyle=dashed,dash=2pt 2pt,linecolor=lightgray](8.660254037844386,12.)(12.138479224601872,9.99184575212343)
\psline[linewidth=0.4pt,linestyle=dashed,dash=2pt 2pt,linecolor=lightgray](12.99160720091437,9.49929208534622)(14.722431864335457,8.5)
\psline[linewidth=0.4pt,linestyle=dashed,dash=2pt 2pt,linecolor=lightgray](14.722431864335457,7.5)(13.854165597947336,8.00129376262745)
\psline[linewidth=0.4pt,linestyle=dashed,dash=2pt 2pt,linecolor=lightgray](11.254750528026433,9.502066752981854)(6.928203230275509,12.)
\psline[linewidth=0.4pt,linestyle=dashed,dash=2pt 2pt,linecolor=lightgray](5.196152422706632,12.)(12.147645281899102,7.9865537264754645)
\psline[linewidth=0.4pt,linestyle=dashed,dash=2pt 2pt,linecolor=lightgray](12.985899331559215,7.502587525254903)(13.856406460551018,7.)
\psline[linewidth=0.4pt,linestyle=dashed,dash=2pt 2pt,linecolor=lightgray](12.990381056766578,6.5)(4.330127018922193,11.5)
\psline[linewidth=0.4pt,linestyle=dashed,dash=2pt 2pt,linecolor=lightgray](3.4641016151377544,11.)(4.326603361283232,10.499588399765926)
\psline[linewidth=0.4pt,linestyle=dashed,dash=2pt 2pt,linecolor=lightgray](6.063109938139454,9.498386923605706)(12.12435565298214,6.)
\psline[linewidth=0.4pt,linestyle=dashed,dash=2pt 2pt,linecolor=lightgray](10.392304845413264,6.)(6.931146430841982,7.998688347875905)
\psline[linewidth=0.4pt,linestyle=dashed,dash=2pt 2pt,linecolor=lightgray](4.333070219488664,9.498688347875905)(2.598076211353316,10.5)
\psline[linewidth=0.4pt,linestyle=dashed,dash=2pt 2pt,linecolor=lightgray](2.598076211353316,9.5)(5.199095623273103,7.998688347875905)
\psline[linewidth=0.4pt,linestyle=dashed,dash=2pt 2pt,linecolor=lightgray](6.065121027057539,7.498688347875906)(8.660254037844386,6.)
\psline[linewidth=0.4pt,linestyle=dashed,dash=2pt 2pt,linecolor=lightgray](6.928203230275509,6.)(2.598076211353316,8.5)
\psline[linewidth=0.4pt,linestyle=dashed,dash=2pt 2pt,linecolor=lightgray](2.598076211353316,7.5)(5.196152422706632,6.)
\psline[linewidth=0.4pt,linestyle=dashed,dash=2pt 2pt,linecolor=lightgray](2.598076211353316,7.5)(2.598076211353316,10.5)
\psline[linewidth=0.4pt,linestyle=dashed,dash=2pt 2pt,linecolor=lightgray](3.4641016151377544,7.)(3.4641016151377544,11.)
\psline[linewidth=0.4pt,linestyle=dashed,dash=2pt 2pt,linecolor=lightgray](4.330127018922193,6.5)(4.333070219488664,7.498688347875906)
\psline[linewidth=0.4pt,linestyle=dashed,dash=2pt 2pt,linecolor=lightgray](4.333070219488664,8.498688347875907)(4.330127018922193,11.5)
\psline[linewidth=0.4pt,linestyle=dashed,dash=2pt 2pt,linecolor=lightgray](5.196152422706632,12.)(5.196152422706632,6.)
\psline[linewidth=0.4pt,linestyle=dashed,dash=2pt 2pt,linecolor=lightgray](6.06217782649107,6.5)(6.06217782649107,11.5)
\psline[linewidth=0.4pt,linestyle=dashed,dash=2pt 2pt,linecolor=lightgray](6.928203230275509,12.)(6.928203230275509,6.)
\psline[linewidth=0.4pt,linestyle=dashed,dash=2pt 2pt,linecolor=lightgray](7.794228634059947,6.5)(7.794228634059947,11.5)
\psline[linewidth=0.4pt,linestyle=dashed,dash=2pt 2pt,linecolor=lightgray](8.660254037844386,12.)(8.660254037844386,6.)
\psline[linewidth=0.4pt,linestyle=dashed,dash=2pt 2pt,linecolor=lightgray](9.526279441628825,6.5)(9.526279441628825,11.5)
\psline[linewidth=0.4pt,linestyle=dashed,dash=2pt 2pt,linecolor=lightgray](10.392304845413264,12.)(10.392304845413264,6.)
\psline[linewidth=0.4pt,linestyle=dashed,dash=2pt 2pt,linecolor=lightgray](11.258330249197702,6.5)(11.277789043465686,7.488765459891273)
\psline[linewidth=0.4pt,linestyle=dashed,dash=2pt 2pt,linecolor=lightgray](11.260782537493263,8.498584170692427)(11.254750528026433,9.502066752981854)
\psline[linewidth=0.4pt,linestyle=dashed,dash=2pt 2pt,linecolor=lightgray](11.254750528026433,9.502066752981854)(11.258330249197702,11.5)
\psline[linewidth=0.4pt,linestyle=dashed,dash=2pt 2pt,linecolor=lightgray](12.12435565298214,12.)(12.12435565298214,6.)
\psline[linewidth=0.4pt,linestyle=dashed,dash=2pt 2pt,linecolor=lightgray](12.990381056766578,6.5)(12.990381056766578,11.5)
\psline[linewidth=0.4pt,linestyle=dashed,dash=2pt 2pt,linecolor=lightgray](13.856406460551018,11.)(13.856406460551018,7.)
\psline[linewidth=0.4pt,linestyle=dashed,dash=2pt 2pt,linecolor=lightgray](14.722431864335457,7.5)(14.722431864335457,10.5)
\psline[linewidth=0.4pt,linestyle=dashed,dash=2pt 2pt,linecolor=lightgray](14.722431864335457,10.5)(12.12435565298214,12.)
\psline[linewidth=1.6pt]{->}(7.78089356293232,8.98416390262379)(9.555058463240279,8.977079469851777)
\psline[linewidth=1.6pt](7.78089356293232,8.98416390262379)(9.500173026150692,8.97729863335183)
\rput[tl](5.1,11.18){$s$}
\rput[tl](11.91,11.18){$s_{m}$}
\begin{scriptsize}
\psdots[dotsize=7pt 0,dotstyle=*,linecolor=red](3.4670448157042264,7.998688347875905)
\psdots[dotsize=7pt 0,dotstyle=*,linecolor=red](4.333070219488664,8.498688347875907)
\psdots[dotsize=7pt 0,dotstyle=*,linecolor=green](4.333070219488664,7.498688347875906)
\psdots[dotsize=7pt 0,dotstyle=*,linecolor=red](5.199095623273103,7.998688347875905)
\psdots[dotsize=7pt 0,dotstyle=*,linecolor=green](6.065121027057539,7.498688347875906)
\psdots[dotsize=7pt 0,dotstyle=*,linecolor=green](6.931146430841982,7.998688347875905)
\psdots[dotsize=7pt 0,dotstyle=*,linecolor=green](4.333070219488664,9.498688347875905)
\psdots[dotsize=7pt 0,dotstyle=*,linecolor=red](5.199095623273103,9.998688347875905)
\psdots[dotsize=7pt 0,dotstyle=*,linecolor=red](4.326603361283232,10.499588399765926)
\psdots[dotsize=7pt 0,dotstyle=*,linecolor=red](5.209770551863357,8.992525174980743)
\psdots[dotsize=7pt 0,dotstyle=*,linecolor=red](6.075795955647811,8.492525174980734)
\psdots[dotsize=7pt 0,dotstyle=*,linecolor=red](10.395370205782715,7.998230213365534)
\psdots[dotsize=7pt 0,dotstyle=*,linecolor=red](11.260782537493263,8.498584170692427)
\psdots[dotsize=7pt 0,dotstyle=*,linecolor=green](11.277789043465686,7.488765459891273)
\psdots[dotsize=7pt 0,dotstyle=*,linecolor=red](12.147645281899102,7.9865537264754645)
\psdots[dotsize=7pt 0,dotstyle=*,linecolor=green](12.985899331559215,7.502587525254903)
\psdots[dotsize=7pt 0,dotstyle=*,linecolor=green](13.854165597947336,8.00129376262745)
\psdots[dotsize=7pt 0,dotstyle=*,linecolor=red](13.01750152033252,8.484341993059656)
\psdots[dotsize=7pt 0,dotstyle=*,linecolor=red](12.126194869203811,8.99893812801932)
\psdots[dotsize=7pt 0,dotstyle=*,linecolor=green](11.254750528026433,9.502066752981854)
\psdots[dotsize=7pt 0,dotstyle=*,linecolor=red](12.138479224601872,9.99184575212343)
\psdots[dotsize=7pt 0,dotstyle=*,linecolor=red](12.99160720091437,9.49929208534622)
\end{scriptsize}
\end{pspicture*}
\caption{Mutation operator applied to the sequence $H^{2}PHP^{2}H^{2}PH^{2}$. The circled node indicate the  position of the mutation point.}
\label{figure5}
\end{figure}

Generally, it consists to modify some components, called genes,  of an existing solution, in order to introduce more diversity into the solutions, it generally applied with low probability. In the  proposed  local search  algorithm (see Algorithm \ref{Algorithm3}),  the  neighbors of a given solution are defined in a similar way with the mutation operator, but the performed movement is  accepted  if it improves the quality of the courant solution. As shown in Figure \ref{figure5}, the new conformation $s_{m}$ is obtained by exciting the current solution $s$ at the mutation point 10, changing the direction from 3 to 6. The energy value of $s_{m}$ is $E(s_{m})=-7$, lower than the energy of $s$, $E(s)=-5$.

\begin{algorithm}[H]
\setstretch{1.48}
\caption{Local Search}
\begin{flushleft}
\algorithmicrequire A feasible  solution $s$, and its associated energy value $E(s)$.\vspace{0.1cm} \\
\algorithmicensure The best found solution $s_{best}$.\\ \vspace{-0.29cm}
\rule{\linewidth}{0.5pt}
\end{flushleft}
\begin{algorithmic}
\State\hspace{-2mm }\textbf{begin}
\State  Initialization:
\State  $E(s_{best})\leftarrow E(s)$;
\State  $s_{best}\leftarrow s$;
\While {the stopping criterion is not checked}
\State $u\leftarrow$ random point of mutation;
\State  $s\leftarrow$ mutation$(s,u)$;
\State Evaluate $s$, i.e., calculate $E(s)$;
\If {$E(s)<E(s_{best})$}
\State $s_{best}\leftarrow s$;
\State $E(s_{best})\leftarrow E(s)$;
\EndIf \State{\textbf{end if}}
\EndWhile  \State{\textbf{end while}}
\State\hspace{-2mm}  \textbf{end}\vspace{4mm}
\end{algorithmic}
\label{Algorithm3}
\end{algorithm}
\subsection{Selection Operator} It is a technique that favorites the best solutions to participate in the reproduction phase, in order to create new solutions with "satisfactory" quality. In this work we will use the roulette wheel selection technique (RWS). In this approach, each individual has a probability of being selected in accordance with his or her performance, so the more individuals are adapted to the problem, the more likely they are to be selected. The probability of selecting a solution $i$ among the $m$ solutions  is given by the following:

$$ p(i)=\dfrac{f(i)}{\dsum_{j=1}^{m}f(j)}\cdot\vspace{2mm}$$

The RWS is an iterative operator, where in each step (after the probability assignment), it withdraws a random value form the range [0,1] (or [0,100], [0,360] depending on the representation of the selection wheel), and then it selects the corresponding individual. Table \ref{table1} presents an example of proportions assignment of four individuals according to their fitness evaluation.

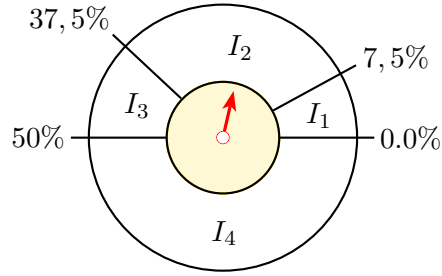
\begin{figure}[H]
\newrgbcolor{ffdxqq}{1. 0.85 0.}
\psset{xunit=1.0cm,yunit=1.0cm,algebraic=true,dimen=middle,dotstyle=o,dotsize=5pt 0,linewidth=0.8pt,arrowsize=3pt 2,arrowinset=0.25}
\begin{pspicture*}(2.23,1.086)(8.52,4.88)
\psaxes[labelFontSize=\scriptstyle,xAxis=true,yAxis=true,Dx=0.5,Dy=0.5,ticksize=-2pt 0,subticks=2]{->}(0,0)(2.23,1.086)(8.52,4.88)
\pscircle[linecolor=ffdxqq,fillcolor=ffdxqq,fillstyle=solid,opacity=0.17](5.,3.){0.74}
\pscircle(5.,3.){1.7640333842602893}
\psline[linewidth=1.2pt,linecolor=red]{->}(5.,3.)(5.14,3.62)
\pscircle(5.,3.){0.74}
\psline(5.66,3.36)(6.75,3.96)
\psline(4.45,3.52)(3.55,4.367)
\psline(5.744,3.00)(6.99,3.)
\psline(4.25,3.00)(3.,3.)
\rput[tl](6.84,4.23){$7,5\%$}
\rput[tl](2.45,4.75){$37,5\%$}
\rput[tl](2.22,3.17){$50\%$}
\rput[tl](7.067,3.16){$0.0\%$}
\rput[tl](6.096,3.46){$I_{1}$}
\rput[tl](5.056,4.36){$I_{2}$}
\rput[tl](3.70,3.59){$I_{3}$}
\rput[tl](4.85,1.90){$I_{4}$}
\begin{scriptsize}
\psdots[linecolor=red](5.,3.)
\end{scriptsize}
\end{pspicture*}
\caption{Roulette wheel selection.}
\label{figure6}
\end{figure}

Figure \ref{figure6} presents the circular representation (i.e., pie chart) of the obtained selection probabilities in Table \ref{table1}.  As it is shown in Figure \ref{figure6}, the spinner indicates to select the chromosome $I_{2}$.
\begin{table}[H]
\centering
\setlength{\tabcolsep}{.5em}
\setlength{\arrayrulewidth}{.05em}
{\renewcommand{\arraystretch}{1.}
\begin{tabular}{ccc}
\hline
Individuals & Fitness: $f_{j}$ & $\%$ of the total sum  \\
\hline
$I_1$ & 3 & 7.5 \\
\hline
$I_2$ & 12 & 30 \\
\hline
$I_3$ & 5 & 12.5 \\
\hline
$I_4$ & 20 & 50 \\
\hline
Total sum & 40 & 100 \\
\hline
\end{tabular}
\caption{Illustrative example of 4 solutions adduced by their fitness evaluations and the corresponding selection proportions.}
}
\label{table1}
\end{table}

\subsection{Generating an offspring  by the guided search strategy}
For the reproduction phase, we  propose an efficient algorithm that guides the search process to  produce an offsprings of "good" quality. This algorithm combines the Local Search algorithm (LS) and the Tabu Search algorithm (TS). The role of the LS algorithm is to improve the quality of the solutions obtained  by the crossover process, and in order to use  the parent's information more effectively and to explore more research space; we propose to use  the tabu search algorithm that memorizes the last points crossover used for some parents selected, see Algorithm \ref{Algorithm4}. The working mechanisms of the suggested approach can be summarized by the following steps:\vspace{3mm}

\begin{description}
  \item[Step1]  Generate $n$ solutions for the initial population.\\
  \item[Step 2]  Evaluate the fitness of each solution.\\
  \item[Step 3]  Create a new population by the following steps:\\
  \begin{description}
    \item[Step3.1]  Select two parents from the $n$ solutions.\\
    \item[Step 3.2]  Apply Algorithm \ref{Algorithm1}, with a probability $p_{c}$.\\
    \item[Step 3.3] Apply Algorithm \ref{Algorithm2}, with a probability $p_{m}$.\\
    \item[Step 3.4]  Replace the previous population, with the current population.\\
  \end{description}
  \item[Step 4]  If the stopping criterion is not checked goto \textbf{Step 2}.\\
\end{description}

\begin{algorithm}[H]
\setstretch{1.4}
\caption{The reproduction algorithm guided by the local search algorithm and the tabu search strategy.}
\begin{flushleft}
\algorithmicrequire Two  selected parents $p_{1}, p_{2}$.\vspace{0.1cm} \\
\algorithmicensure Two individuals produced from parents, i.e., offsprings.\\ \vspace{-0.29cm}
\rule{\linewidth}{0.5pt}
\end{flushleft}
\begin{algorithmic}[!h]
\State\hspace{-2mm }\textbf{Begin}
\State  Initialization:
\State $E_{1} \leftarrow E(p_{1})$;
\State  $offspring^{*}_{1}\leftarrow p_{1}$;
\State  $E_{2}\leftarrow E(p_{2})$;
\State $offspring^{*}_{2}\leftarrow p_{2}$;
\State $T\longleftarrow \emptyset$;   // the tabu list
\State  $K leftarrow 0$;
\While {the stopping criterion is not checked}
\State $K\leftarrow K+1 $;
\State $u\leftarrow$ random point of crossover;
\If { $u\notin T$ }
\State $(offspring_{1}, offspring_{2})\leftarrow$ crossover $(p_{1},p_{2})$;
\State $u_{1} \leftarrow$ random $[0,1];$
\If {$ u_{1} \leq p_{m}$}
\State $offspring_{1}\leftarrow$ local search $(offspring_{1})$;
\State $offspring_{2}\leftarrow$ local search $(offspring_{2})$;
\EndIf \State{\textbf{end if}}
\If{ $(E(offspring_{1})<E_{1})$}
\State $ offspring^{*}_{1}\leftarrow offspring_{1}$;
\State $E_{1}\leftarrow E(offspring_{1})$;
\EndIf \State{\textbf{end if}}
\If { $ (E(offspring_{2})<E_{2})$}
\State $ offspring^{*}_{2}\leftarrow offspring_{2}$;
\State $ E_{2}\leftarrow E(offspring_{2}) $;
\EndIf \State{\textbf{end if}}
\EndIf \State{\textbf{end if}}
\State \hspace{0.4cm} $ T\longleftarrow T\cup \{ u \}$;
\EndWhile \State{\textbf{end while}}
\State \Return {$(offspring^{*}_{1},offspring^{*}_{2})$;}
\State\hspace{-2mm}  \textbf{end}\vspace{4mm}
\end{algorithmic}
\label{Algorithm4}
\end{algorithm}
\section{Experimental Results}
The aim of this section is to assess the performance of the suggested approach. For the following experimental study, we used a several benchmark instances (i.e., protein sequences) presented in the H-P model of different sizes \cite{dandekar_folding_1994, unger_genetic_1993}. Furthermore, we have selected the most studied in the literature instances to conduct the forthcoming experiments. Table \ref{table2} is composed of 3 columns, the first one represents the number of the protein sequence, the second represents  the length of the protein sequence and the third represents the protein sequence in the H-P model, where the symbol $(.)^{i}$ means $i$ fold repetitions of the respective subsequence in the brackets. For example, $(PH)^2$ is the simplified form of the sequence $PHPH$. The experimental results showed in Table \ref{table3} represents  a comparison between  the best results obtained by the proposed  approach, and the above stated algorithms used to solve the PSP in the 2D triangular lattice model, namely HGA \cite{hoque_hybrid_2006}, TS \cite{bockenhauer_local_2008}, ERS-GA  \cite{su_efficient_2010}, HHGA \cite{su_efficient_2010}, IMOG \cite{yang_protein_2018} for each instance given in Table \ref{table2}. We show that all the approaches can provide an optimal confirmation when the length of the  sequence is less than 36. However, the results obtained by our approach are better than the other approaches for the sequences 4,\ 6 and 7 as shown in Table \ref{table3} (see the respective conformation in Figure \ref{figure9}). However, The best prediction obtained by the algorithms GALSTS, TS , HHGA and  IMOG  are  better  than the HGA algorithm, ERS-GA and SGA  for all sequences used for this experimental study.
\begin{table}[H]
\centering
\setlength{\tabcolsep}{.5em}
\setlength{\arrayrulewidth}{.05em}
{\renewcommand{\arraystretch}{1.45}
\begin{footnotesize}
\begin{tabular}{ccl}
\hline
Seq. & Length & Protein sequence in the H-P model  \\
\hline
$1$ & $20$ & $(HP)^{2}PH(HP)^{2}(PH)^{2}HP(PH)^{2}$ \\
\hline
$2$ & $24$ & $H^{2}P^{2}(HP^{2})^{6}H^{2}$ \\
\hline
$3$ & $25$ & $P^{2}HP^{2}(H^{2}P^{4})^{3}H^{2}$ \\
\hline
$4$ & $36$ & $P(P^{2}H^{2})^{2}P^{5}H^{5}(H^{2}P^{2})^{2}P^{2}H(HP^{2})^{2}$ \\
\hline
$5$ & $40$ & $P^{2}H(P^{2}H^{2})^{2}P^{5}H^{10}P^{6}(H^{2}P^{2})^{2}HP^{2}H^{5}$ \\
\hline
$6$ & $50$ & $H^{2}(PH)^{3}PH^{4}PH(P^{3}H)^{2}P^{4}(HP^{3})^{2}HPH^{4}(PH)^{3}PH^{2}$ \\
\hline
$7$ & $60$ & $P(PH^{3})^{2}H^{5}P^{3}H^{10}PHP^{3}H^{12}P^{4}H^{6}PH^{2}PHP$ \\
\hline
$8$ & $64$ & $H^{12}(PH)^{2}((P^{2}H^{2})^{2}P^{2}H)^{3}(PH)^{2}H^{11}$ \\
\hline
$9$ & $85$ & $H^{4}P^{4}H^{12}P^{6}(H^{12}P^{3})^{3}HP^{2}(H^{2}P^{2})^{2}HPH$ \\
\hline
$10$ & $100$ & $P^{3}H^{2}P^{2}H^{4}P^{2}H^{3}(PH^{2})^{3}H^{2}P^{8}H^{6}P^{2}H^{6}P^{9}HPH^{2}PH^{11}P^{2}H^{3}PH^{2}PHP^{2}HPH^{3}P^{6}H^{3}$ \\
\hline

\end{tabular}
\caption{The used benchmark instances in the H-P model.}
\label{table2}
\end{footnotesize}}
\end{table}

The experiment results, in Tables \ref{table3} and \ref{table4}, represent a comparison between the best results obtained by GALSTS, with some existing results based on several large  benchmarks found in the literature, solved by HGA and SGA \cite{hoque_hybrid_2006}, TS \cite{bockenhauer_local_2008}, ERS-GA \cite{su_efficient_2010}, HHGA \cite{su_efficient_2010}, and IMOG \cite{yang_protein_2018} approaches. Clearly, the number of possible conformations increases exponentially when the size of the instance increases. According to their energy values, all approaches could provide an optimal confirmation when the size of the instance is less than 36. But, as it is shown in \mbox{Tables \ref{table3}} and \ref{table4}, the best conformations obtained by GALSTS are better than all the cited approaches bellow, based on the instances of size up to 36. This demonstrates the ability of GALSTS to explore the search space more effectively comparing with the other approaches.

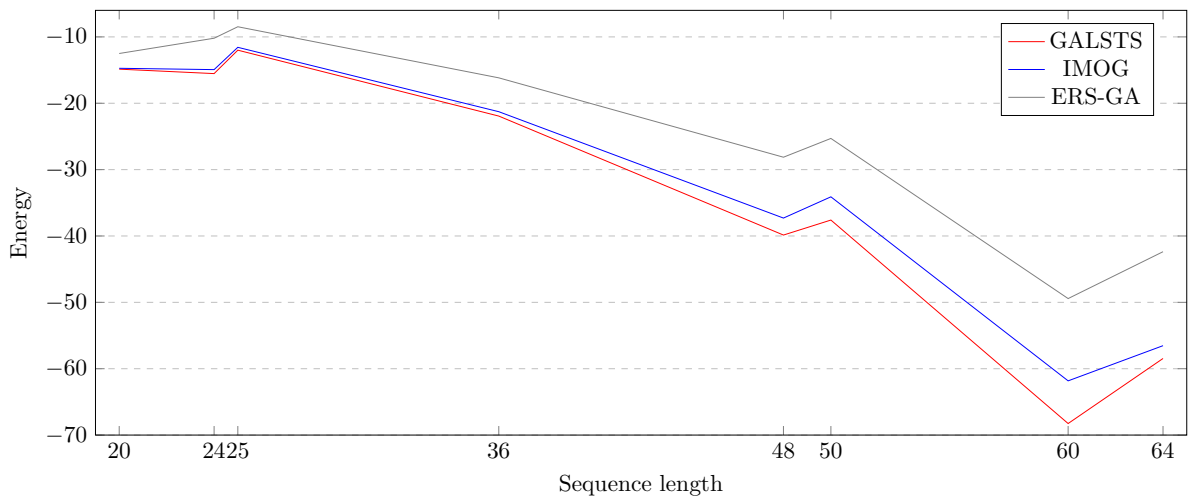
\begin{figure}[h!]
\pgfplotsset{width=8cm}
\begin{tikzpicture}[transform shape, scale = 0.8]
\hspace{-.7cm}
\begin{axis}[
    title={},
    xlabel={Sequence length},
    ylabel={Energy},
    xmin=19, xmax=65,
    ymin=-70, ymax=-6,
    x=3.9mm,
    y=1.1mm,
    xtick=data,
    ytick={-70, -60, ...,-10},
    legend pos=north east,
    ymajorgrids=true,
    grid style=dashed,enlargelimits=false
]

\addplot[
 color = red,
    ]
    coordinates {
(20, -14.86) (24 , -15.53) (25 , -12) (36 ,-21.93) (48,-39.86) (50 ,-37.6) (60 ,-68.26) (64 ,-58.46)
    };
    \addlegendentry{GALSTS}

  \addplot[
 color = blue,
    ]
    coordinates {
(20, -14.73) (24 , -14.93) (25 , -11.57) (36 ,-21.27 ) (48,-37.30 ) (50 ,-34.10 ) (60 ,-61.83) (64 ,-56.53)
    };
    \addlegendentry{IMOG}
 \addplot[
 color = gray,
    ]
    coordinates {
(20, -12.50) (24 , -10.20) (25 , -8.47 ) (36 ,-16.17) (48,-28.13) (50 ,-25.30) (60 ,-49.43) (64 ,-42.37)
    };
    \addlegendentry{ERS-GA}
\end{axis}
\end{tikzpicture}
\caption{Illustration of the comparison results regarding the mean energy obtained using GALSTS against other algorithms.}
\label{figure7}
\end{figure}

\begin{figure}[h!]
\pgfplotsset{width=8cm}
\begin{tikzpicture}[transform shape, scale = 0.8]
\hspace{-.7cm}
\begin{axis}[
    title={},
    xlabel={Sequence length},
    ylabel={Energy},
    xmin=19, xmax=65,
    ymin=-73, ymax=-10,
    x=3.9mm,
    y=1.1mm,
    xtick=data,
    ytick={-70, -60, ...,-10},
    legend pos=north east,
    ymajorgrids=true,
    grid style=dashed,enlargelimits=false
]

\addplot[
 color = red,
    ]
    coordinates {
(20, -15) (24 , -17) (25 , -12) (36 ,-24) (48,-43) (50 ,-40) (60 ,-70) (64 ,-67)
    };
    \addlegendentry{GALSTS}

  \addplot[
 color = blue,
    ]
    coordinates {
(20, -15) (24 , -17) (25 , -12) (36 ,-23 ) (48,-41 ) (50 ,-38 ) (60 ,-66) (64 ,-63)
    };
    \addlegendentry{IMOG}
 \addplot[
 color = gray,
    ]
    coordinates {
(20, -15) (24 , -13) (25 , -12 ) (36 ,-20) (48,-32) (50 ,-30) (60 ,-55) (64 ,-47)
    };
    \addlegendentry{ERS-GA}
\end{axis}
\end{tikzpicture}
\caption{Illustration of the comparison results regarding the best predictions obtained using GALSTS against other algorithms.}
\label{figure8}
\end{figure}
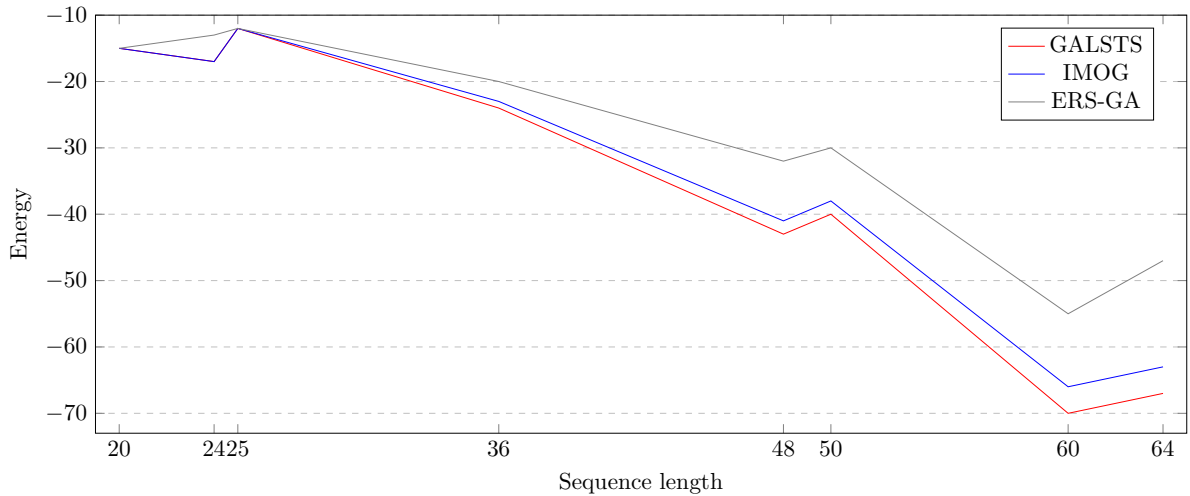

\begin{figure}[H]
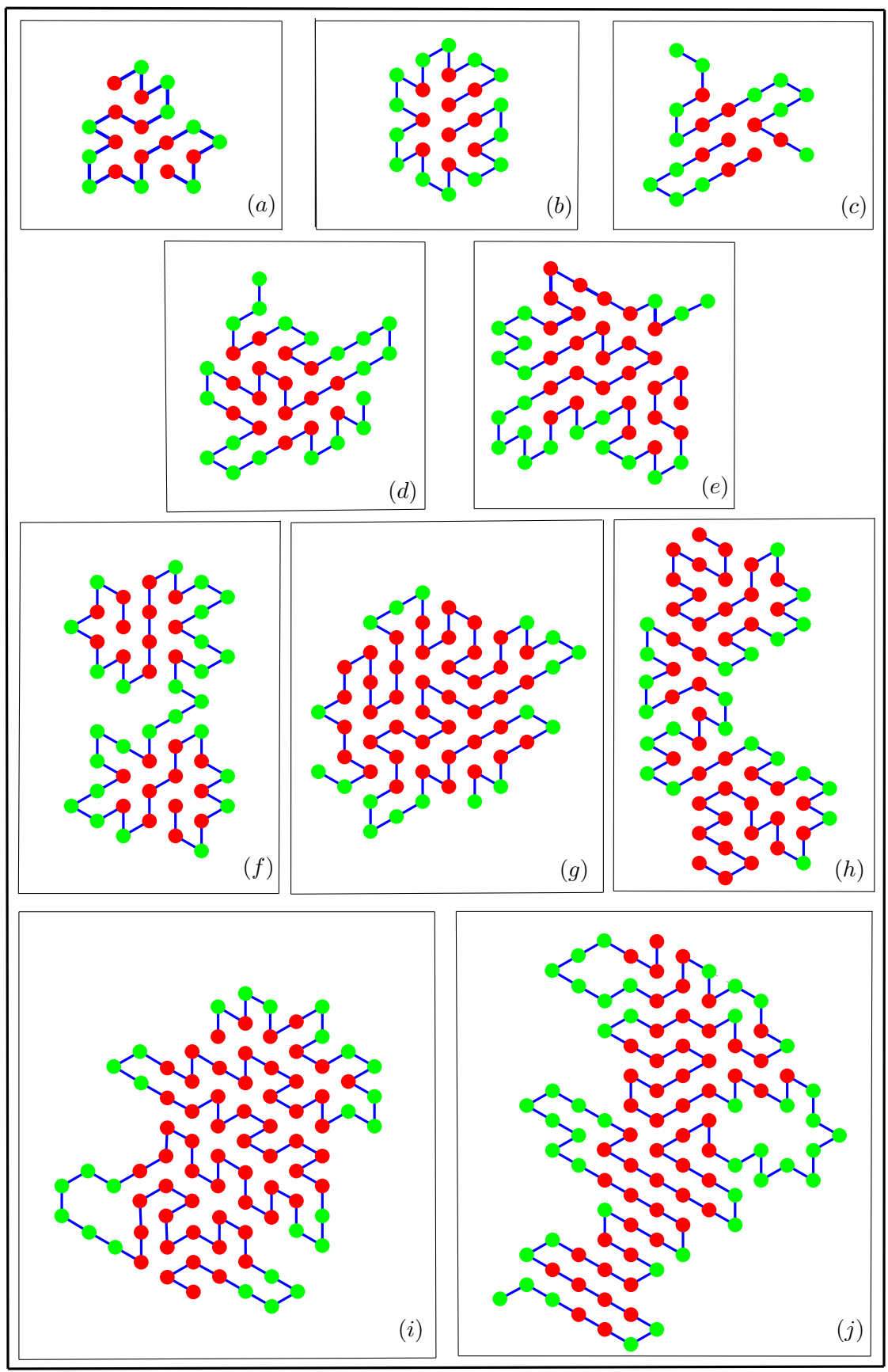

$$
$$\vspace{-0.5cm}
\caption{$(a)$ to $(j)$ Results of the best conformation structure of ten protein sequences.}
\label{figure9}
\end{figure}

\begin{table}[H]
%\begin{figure}[H]
\centering
\setlength{\tabcolsep}{0.67em}
\setlength{\arrayrulewidth}{.05em}
{\renewcommand{\arraystretch}{1.3}
\begin{footnotesize}
\begin{tabular}{cccccccccc}
\hline
Seq. & Length & SGA& HGA& TS & ERS-GA & HHGA & IMOG & GALSTS & Conformation\\ \hline
\hline
1 & 20 & -11 & \textbf{-15} & \textbf{-15} & \textbf{-15} & \textbf{-15} & \textbf{-15} & \textbf{-15}& Fig\ref{figure7}$(a)$\\ \hline
2 & 24 & -10 & -13 & \textbf{-17} & -13 & \textbf{-17} & \textbf{-17} & \textbf{-17}&  Fig.\ref{figure7}$(b)$\\ \hline
3 & 25 & -10 & -10 & \textbf{-12} & -12 & \textbf{-12} & \textbf{-12} & \textbf{-12}& Fig.\ref{figure7}$(c)$\\ \hline
4 & 36 & -16 & -19 & \textbf{-24} & -20 & -23 & \textbf{-24} & \textbf{-24}& Fig.\ref{figure7}$(d)$\\ \hline
5 & 48 & -26 & -32 & -40 & -32 & -41 & -40 & \textbf{-43}& Fig.\ref{figure7}$(e)$\\ \hline
6 & 50 & -21 & -23 & NA & -30 & -38 & \textbf{-40} & \textbf{-40}& Fig.\ref{figure7}$(f)$\\ \hline
7 & 60 & -40 & -46 & \textbf{-70} & -55 & -66 & -67 & \textbf{-70}& Fig.\ref{figure7}$(g)$\\ \hline
8 & 64 & -33 & -46 & \textbf{-50} & -47 & -63 & -63 & \textbf{-67}& Fig.\ref{figure7}$(h)$\\ \hline
9 & 85 &  NA   &  NA   &   NA      &  NA    &  NA   &  NA   &   \textbf{-98} & Fig.\ref{figure7}$(i)$\\ \hline
10 & 100 &  NA   &  NA   &   NA      &  NA    &  NA   &  NA   &   \textbf{-87}& Fig.\ref{figure7}$(j)$\\ \hline
 \multicolumn{10}{l}{Values in bold indicate the best obtained evaluation for the correspondent instance.}\\
  \multicolumn{10}{l}{NA refers to not available data in literature.}
\end{tabular}
\caption{The best conformations obtained by GALSTS compared with other algorithms for 10 H-P sequences in 2D triangular lattice model.}
\label{table3}
\end{footnotesize}}
\end{table}
Instances larger than 64 are not covered in the literature for the 2D triangular lattice model. However, they were processed for the rectangular model \cite{zhao_advances_2008}. According to the obtained results in Table \ref{table3}, a strong improvement in energy appears clearly compared with the triangular model. Table \ref{table4}, graphically presented  in Figure \ref{figure7} and Figure \ref{figure8}, show a performance comparison  on the  stability of our approach and three  other  algorithms  HHGA, IMOG and  ERS-GA, such that the efficiency of the algorithms is measured by the  best and means results  in 30 independent runs for each sequence. We  show  that for the  most  instances, the proposed approach is able to find the best optimal  solutions and achieves a better average solution quality than other algorithms. the average solution quality is very encouraging.
\begin{table}[H]
%\begin{figure}[H]
\centering
\setlength{\tabcolsep}{0.67em}
\setlength{\arrayrulewidth}{.05em}
{\renewcommand{\arraystretch}{1.3}
\begin{footnotesize}
\begin{tabular}{ccccccccccccc}
\hline
 &  &\multicolumn{2}{c}{ERS-GA }   &  & \multicolumn{2}{c}{HHGA}  &  & \multicolumn{2}{c}{IMOG} &  &  \multicolumn{2}{c}{GALSTS}  \\
\cline{3-4}\cline{6-7}\cline{9-10}\cline{12-13}
Seq. & Length & Best  & Mean &  & Best & Mean  &  & Best  & Mean &  & Best & Mean \\
\hline
1 & 20 & \textbf{-15} & -12.50 &  & \textbf{-15} & -14.73 &  & \textbf{-15} & -14.73 &  & \textbf{-15} & \textbf{-14.86} \\
2 & 24 & -13 & -10.20 &  & \textbf{-17} & -14.93 &  & \textbf{-17} & -14.93 & &\textbf{-17} & \textbf{-15.53} \\
3 & 25 & 12 & -8.47 &  & \textbf{-12} & - 11.57 &  & \textbf{-12} & -11.57 &  & \textbf{-12} & \textbf{-12} \\
4 & 36 & -20 & -16.17 &  & -23 & -21.27 &  & -23 & -21.27 &  & \textbf{-24} & \textbf{-21.93} \\
5 & 48 & -32 & -28.13 &  & -41 & - 37.30 &  & -41 & -37.30 & &\textbf{-43} & \textbf{-39.86} \\
6 & 50 & -30 & -25.30 &  & -38 & -34.10 &  & -38 & -34.10 &  & \textbf{-40} & \textbf{-37.6} \\
7 & 60 & -55 & -49.43 &  & -66 & - 61.83 &  & -66 & -61.83 &  & \textbf{-70} & \textbf{-68.26} \\
8 & 64 & -47 & -42.37 &  & -63 & - 56.53 &  & -63 & -56.53 &  & \textbf{-67} & \textbf{-58.46} \\\hline
 \multicolumn{11}{l}{Values in bold indicate the best obtained evaluation for the correspondent instance.}

\end{tabular}
\vspace{0.2cm}
\caption{ A comparative study on the stability and the best prediction of the GALSTS with other algorithms.}
\label{table4}
\end{footnotesize}}
\end{table}

Table 5 resumes the obtained results for 30 independent runs per each of the above stated instances. The aim of this experiment is to compare the suggested algorithm GALSTS against two competing algorithms ERS-GA and SGA. The results are adduced according to the best and worst overall evaluation and their corresponding deviation from the best known value (BKV). The proposed algorithm GALSTS is shown to be more effective than the other competing algorithms; even when comparing its worst produced conformation to their best ones, with a sole exception of the first tested sequence, where it shows a slight difference. However, when increasing the size of the instance, it is clear that the suggested algorithm is incrementally taking advantage over the competing algorithms, even when comparing its worst solution to their best ones. Furthermore, the suggested algorithm is shown to be able to attain good quality conformations or even optimal, with a sole exception the sixth tested instance, where it shows a one unit deviation of the best known evaluation.
\begin{table}[H]
%\begin{figure}[H]
\centering
\setlength{\tabcolsep}{0.7em}
\setlength{\arrayrulewidth}{.05em}
{\renewcommand{\arraystretch}{1.3}
\begin{footnotesize}
\begin{tabular}{ccccccccc}
\hline
 &  &  & \multicolumn{2}{c}{GALSTS} &  & ERS-GA &  & SGA\\
\cline{4-5}\cline{7-7}\cline{9-9}
Seq. & length & $E^{*}$ & \multicolumn{1}{c}{Best (dev. BKV)} & \multicolumn{1}{c}{Worst } & \multicolumn{1}{c}{} & \multicolumn{1}{c}{Best (dev. BKV)} & \multicolumn{1}{c}{} & Best (dev. BKV) \\
\hline
1 & 20 & -15 & \multicolumn{1}{c}{\textbf{-15} (\textbf{00})} & \multicolumn{1}{c}{-14} & \multicolumn{1}{c}{} & \multicolumn{1}{c}{\textbf{-15} (\textbf{00})} & \multicolumn{1}{c}{} & -11 (04) \\
2 & 24 & -17 & \multicolumn{1}{c}{\textbf{-17} (\textbf{00})} & \multicolumn{1}{c}{-15} & \multicolumn{1}{c}{} & \multicolumn{1}{c}{-13 (04)} & \multicolumn{1}{c}{} & -10 (07) \\
3 & 25 & -12 & \multicolumn{1}{c}{\textbf{-12} (\textbf{00})} & \multicolumn{1}{c}{-12} & \multicolumn{1}{c}{} & \multicolumn{1}{c}{\textbf{-12} (\textbf{00})} & \multicolumn{1}{c}{} & -10 (02)\\
4 & 36 & -24 & \multicolumn{1}{c}{\textbf{-24} (\textbf{00})} & \multicolumn{1}{c}{-21} & \multicolumn{1}{c}{} & \multicolumn{1}{c}{-20 (04)} & \multicolumn{1}{c}{} & -16 (08)\\
5 & 48 & -43 & \multicolumn{1}{c}{\textbf{-43} (\textbf{00})} & \multicolumn{1}{c}{-38} & \multicolumn{1}{c}{} & \multicolumn{1}{c}{-32 (11)} & \multicolumn{1}{c}{} & -26 (17)\\
6 & 50 & -41 & \multicolumn{1}{c}{\textbf{-40} (\textbf{01})} & \multicolumn{1}{c}{-36} & \multicolumn{1}{c}{} & \multicolumn{1}{c}{-30 (11)} & \multicolumn{1}{c}{} & -21 (20)\\
7 & 60 & - & \multicolumn{1}{c}{\textbf{-70} (-)} & \multicolumn{1}{c}{-65} & \multicolumn{1}{c}{} &\multicolumn{1}{c}{-55 (-)} & \multicolumn{1}{c}{} & -40 (-) \\
8 & 64 & - & \multicolumn{1}{c}{\textbf{-67} (-)} & \multicolumn{1}{c}{-56} & \multicolumn{1}{c}{} & \multicolumn{1}{c}{-47 (-)} & \multicolumn{1}{c}{} & -33 (-) \\
\hline
 \multicolumn{9}{l}{Values in bold indicate the best obtained evaluation for the correspondent instance.}
\end{tabular}
\vspace{0.2cm}
\caption{ Comparison between the results, the best and worst evaluations obtained by  GALSTS and the best results of SGA and ERS-GA. $E^{*}$ is the best energy value.}
\label{table5}
\end{footnotesize}}
\end{table}

%%%%%%%%%%%%%%%%%%%%%%%%%%%%%%%%%%%%
% Graphes %%%%%%%%%%%%%%%%%%%%%%%%%%
%%%%%%%%%%%%%%%%%%%%%%%%%%

\section{ Conclusions and Perspectives}
This paper presents an efficient approach called GALSTS for solving the protein structure prediction in 2D triangular model using the simplified hydrophobic-polar model. An initialization algorithm was proposed that allows to generate only valid conformations  for the initial population of GALSTS. This algorithm eliminates cyclic movements during the construction of solutions. GALSTS consists in using Tabu and Local Search algorithm  to explore the search space  handling more efficiently. This approach  allows to use the information provided by the selected parents to produce solutions with good quality. From our experimental results, GALSTS was   able to  find the best known solutions and it is more effective for the stability results  than other existing algorithms. In terms of future scope of applications, GALSTS can be used to solve the PSP problem in the 3D cubic and 3D triangular models, it can also be used to solve other optimization problems in the combinatorial optimization framework.
\section{Acknowledgment} The authors thank N. Kantour for his generous assistance. 
\newpage
\bibliographystyle{acm}
\bibliography{bibo}

%\end{bibliography}
\end{document}